\documentclass[letterpaper]{article} 
\usepackage[preprint]{narxiv2026}  
\usepackage[hyphens]{url}  
\usepackage{graphicx} 
\usepackage{natbib}  
\usepackage{caption} 
\usepackage{algorithm}
\usepackage{algorithmic}
\usepackage{amssymb}
\usepackage{tabularx}
\usepackage{booktabs}
\usepackage{array}
\usepackage{amsmath}
\usepackage{multirow}
\usepackage{graphicx}
\usepackage{makecell}
\usepackage{xcolor}

\usepackage{newfloat}
\usepackage{listings}
\DeclareCaptionStyle{ruled}{labelfont=normalfont,labelsep=colon,strut=off} 
\floatstyle{ruled}
\newfloat{listing}{tb}{lst}{}
\floatname{listing}{Listing}

\usepackage{booktabs}

\def\copyright@text{%
\textit{Preprint. Under review.}
}

\title{Training-Free Entity-Level Few-Shot Segmentation of Remote Sensing Images with Advection Refinement}
\author{
    Xueting Bai,
    Huan Ni\corresponding
}
\affiliations{
    School of Remote Sensing and Geomatics Engineering, Nanjing University of Information Science and Technology, \\Nanjing 210044, China.\\
    nih@nuist.edu.cn
}

\begin{document}

\maketitle

\begin{abstract}
Existing cross-domain few-shot segmentation approaches suffer from high training costs due to source-domain episodic training and pixel-wise dense prediction, while often producing fragmented and noisy predictions in remote sensing images. To overcome these issues, we propose ELFSS-AR, a training-free entity-level few-shot segmentation method with advection refinement. Specifically, ELFSS-AR first leverages SAM3's generic geometric priors to generate category-agnostic entity primitives, and reformulates few-shot inference from pixel-level prediction to entity-level reasoning. Subsequently, foreground and background prototypes are constructed and combined with dense textual semantic responses from SAM3 to build a multi-modal semantic potential field. An advection equation-based semantic refinement mechanism is then introduced to propagate category-aware information across both feature and multi-source response spaces, further enhancing semantic continuity and suppressing local texture noise. Extensive experiments on multiple remote sensing datasets demonstrate that ELFSS-AR effectively mitigates domain shift and local noise, substantially improving SAM3's adaptation capability for remote sensing few-shot segmentation without additional training. The code is publicly available at https://github.com/yu-ni1989/ELFSS-AR. 
\end{abstract}

\section{Introduction}
\label{sec:intro}

Few-shot segmentation (FSS) aims to predict dense object masks for query images using one or a few support images with pixel-level annotations~\cite{Shaban-2017}. Existing studies mainly follow the episodic training paradigm~\cite{Snell-2017}. Considerable efforts have been devoted to improving segmentation performance from various perspectives, including prototype refinement~\cite{PANet-2019,PFENet-2022,PFENet-2024,SSP-2022}, matching calibration~\cite{Peng-2023,DSV-LFS-2025,Zou-2025}, classifier adaptation~\cite{Lu-2023}, and category bias suppression~\cite{Lang-2022}. Recently, foundation segmentation models, such as SAM~\cite{SAM-2023}, which can generate category-agnostic entity primitives, have also been introduced into FSS to enhance object localization~\cite{Xu-2025,Shi-2026}. However, when significant feature distribution discrepancies exist between the source and target domains, conventional FSS methods often exhibit limited cross-domain generalization ability. To address this issue, Cross-Domain Few-Shot Segmentation (CDFSS) further investigates how a small number of target-domain support samples can alleviate domain shift~\cite{CDFSS-Lei-2022}. Existing approaches mainly focus on target-domain fine-tuning~\cite{CDFSS-Nie2024,CDFSS-Herzog2024}, domain-invariant feature learning~\cite{CDFSS-Su2024,CDFSS-Tong-2025}, matching calibration~\cite{CDFSS-Li-2025,CDFSS-Liu-2025}, and prompt generation based on foundation models~\cite{CDFSS-He-2024}. Although these methods have achieved promising progress, most of them still rely on supervised training in the source domain and incur substantial training costs. Moreover, these methods generally perform pixel-wise prediction, which often leads to unstable matching and inconsistent intra-region responses under complex scale variations, spectral discrepancies, and high-frequency texture interference commonly encountered in remote sensing imagery.

\begin{table*}[t]
\centering
\small
\setlength{\tabcolsep}{4pt}
\renewcommand{\arraystretch}{1.15}
\caption{Comparison between ELFSS-AR and representative FSS, CDFSS, foundation model-based FSS, training-free OVSS, and support-assisted OVS methods. Symbols $\checkmark$, $\triangle$, and $\times$ denote fully supported, partially supported, and unsupported, respectively.}
\label{tab:comparison}
\begin{tabularx}{\textwidth}{l>{\raggedright\arraybackslash}Xcccc}
\toprule
\textbf{Framework} & \textbf{Representative Methods} & \shortstack{\textbf{Training-}\\\textbf{Free}} & \shortstack{\textbf{Entity-Level}\\\textbf{Inference}} & \shortstack{\textbf{Physics-Inspired}\\\textbf{Refinement}} & \shortstack{\textbf{Multimodal}\\\textbf{Prompting}} \\
\midrule
FSS & PANet~\cite{PANet-2019}, PFENet++~\cite{PFENet-2024} & $\times$ & $\times$ & $\times$ & $\times$ \\
CDFSS & DR-Adapter~\cite{CDFSS-Su2024}, DATO~\cite{CDFSS-Li-2025} & $\times$ & $\times$ & $\times$ & $\times$ \\
Foundation model-based FSS & APSeg~\cite{CDFSS-He-2024}, BPNet~\cite{Shi-2026} & $\times$ & $\triangle$ & $\times$ & $\triangle$ \\
Training-free OVSS & SCLIP~\cite{SCLIP-2025}, ProxyCLIP~\cite{ProxyCLIP-2025} & $\checkmark$ & $\times$ & $\times$ & $\checkmark$ \\
Support-assisted OVS & ConceptBank~\cite{conceptbank_2026} & $\checkmark$ & $\times$ & $\times$ & $\checkmark$ \\
ELFSS-AR (Ours) & ELFSS-AR (Ours) & $\checkmark$ & $\checkmark$ & $\checkmark$ & $\checkmark$ \\
\bottomrule
\end{tabularx}
\end{table*}

Benefiting from large-scale pretraining on massive vision-language corpora, multimodal foundation models possess category-agnostic region modeling, prompt interaction, and zero-shot transfer capabilities~\cite{SAM-2023,SAM2-2025,SAM3-2026}. Building upon these advantages, recent multimodal frameworks have further explored multimodal prompt interaction~\cite{Zou-2023b,Liu-2025}, unified modeling across various segmentation tasks~\cite{Zhang-2023,Li-2025}, and pixel-level prediction~\cite{LISA-2024,GLaMM-2024,PixelLM-2024}. Among them, training-free open-vocabulary semantic segmentation (OVSS) methods directly exploit frozen vision-language models for dense prediction through local vision-language alignment~\cite{Zhou-2022}, self-attention and cross-modal propagation~\cite{SCLIP-2025,ClearCLIP-2025,Shao-2025,Hajimiri-2025}, multi-level feature fusion~\cite{ResCLIP-2025,Shi-2025}, external visual priors~\cite{ProxyCLIP-2025}, feature restoration~\cite{SegEarth-OV-2025}, and attention reconstruction~\cite{ReAttnCLIP-2026}, thereby improving remote sensing domain adaptation. These approaches provide a promising low-cost solution for domain transfer without additional model training. Nevertheless, since these foundation models are primarily pretrained on natural images and generic vision-language corpora, they lack dedicated modeling for the unique characteristics of remote sensing imagery, including bird's-eye viewpoints, significant scale variations, and complex spectral-textural patterns. Moreover, land-cover classes in remote sensing often exhibit highly domain-specific semantics, particularly in fine-grained land-cover and land-use classification tasks. For example, the industrial area class often encompasses a variety of objects, including internal roads, buildings, and green spaces. Categories such as ponds, fishponds, paddy fields, rivers, and ditches are also easily confused. Consequently, SAM3~\cite{SAM3-2026} achieves very low accuracy on fine-grained land-cover and land-use datasets such as the GID-15 (Gaofen Image Dataset with 15 categories)~\cite{TONG-2020} and MiniFrance~\cite{MiniFrance-2020} datasets.

\begin{figure}[t]
\centering
\includegraphics[width=0.47\textwidth]{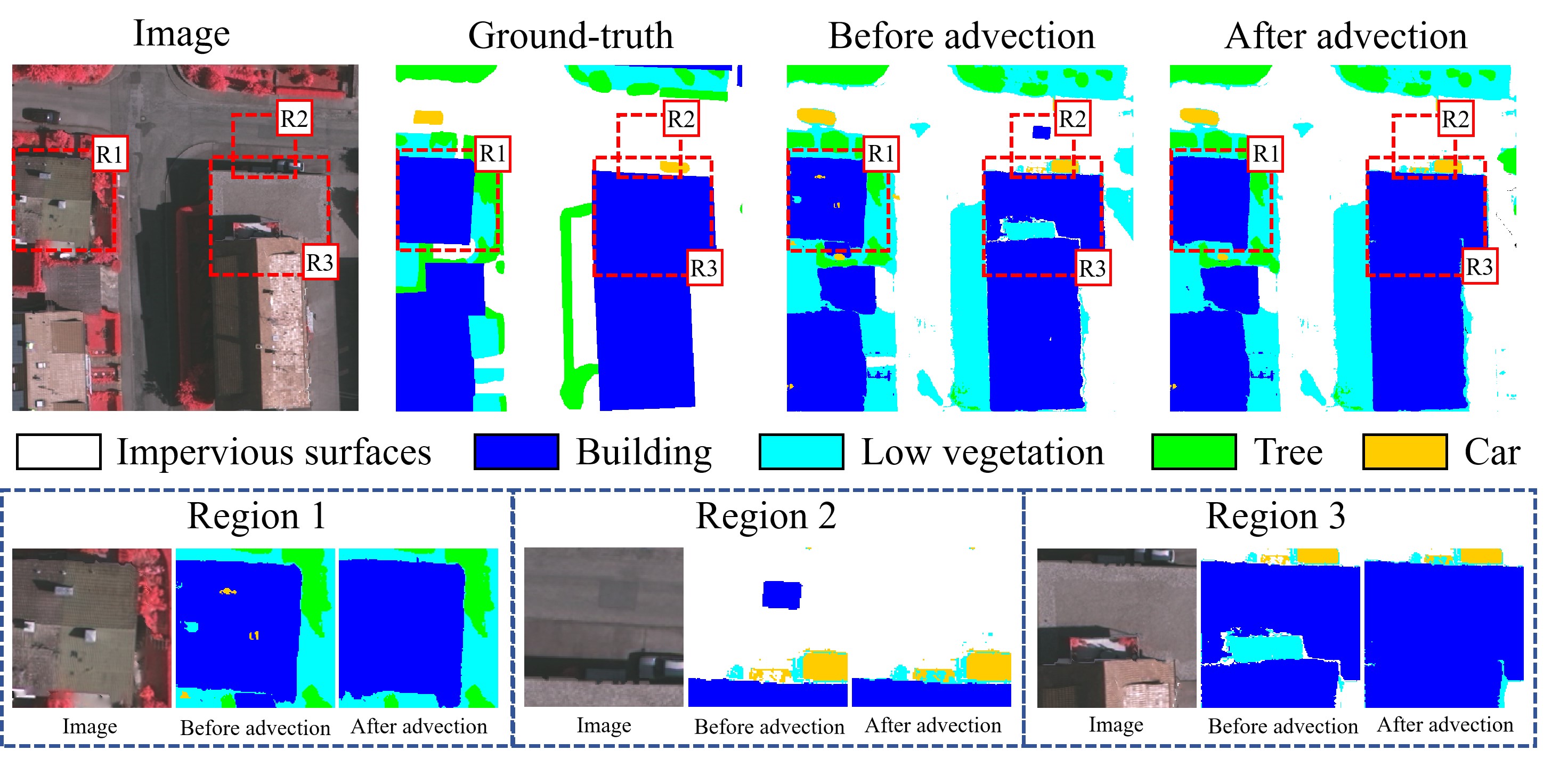}
\caption{Illustration of the effect of Semantic Advection refinement. }
\label{fig:advection}
\end{figure}

To address these issues, we build on the CDFSS paradigm and leverage SAM3's strengths in class-agnostic entity primitive generation and text modeling. This approach improves the model's ability to adapt to complex remote-sensing data distributions while extending CDFSS from pixel-level to entity-level prediction. Experiments show that it substantially improves accuracy on remote-sensing land-cover and land-use datasets (see the Ablation Study section). However, as shown in Fig.~\ref{fig:advection}, even when the target region is correctly predicted overall, local noise or holes still occur because of variations in texture, shadows, and spectral responses. To further address this issue, inspired by the physical process of advection, we propose a semantic advection refinement mechanism. It propagates high-confidence semantic responses through a semantic potential field into low-confidence regions, thereby filling local holes, removing isolated noisy responses, and enhancing semantic continuity within regions. A theoretical analysis of the mechanism's effectiveness is provided in \textit{Appendix}. Finally, we integrate these innovations to form the ELFSS-AR, a training-free framework for entity-level FSS of remote-sensing images with advection refinement. Table~\ref{tab:comparison} summarizes the distinctive features of ELFSS-AR compared with existing methods. 

\section{Methods}
\label{sec:methods}

\subsection{Overall Framework}
\label{sec:overall}

\begin{figure*}[t]
\centering
\includegraphics[width=1.0\textwidth]{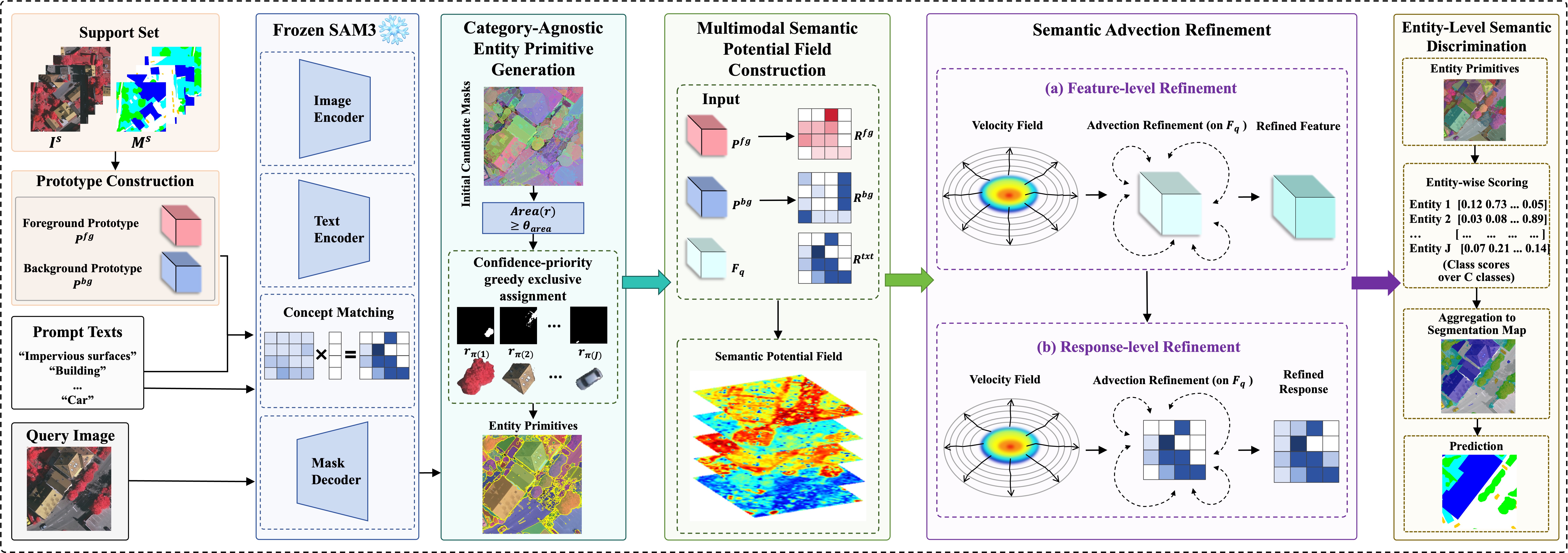} 
\caption{Overview of the proposed Training-Free Entity-Level Few-Shot Segmentation framework with Advection Refinement (ELFSS-AR). The proposed framework first generates category-agnostic entity primitives, and then integrates support prototypes with textual semantics to construct a multimodal semantic potential field. Guided by the advection equation, the query features and multi-source similarity responses are progressively refined, followed by entity-level classification to produce the final segmentation results.}
\label{fig_overall}
\end{figure*}

The overall architecture of ELFSS-AR is illustrated in Fig.~\ref{fig_overall}. ELFSS-AR adopts SAM3 as the feature extraction backbone and first generates a set of category-agnostic entity primitives, extending the conventional pixel-wise inference paradigm of FSS and CDFSS to entity-level inference. Subsequently, visual and textual features are jointly integrated to construct a multimodal semantic potential field. Based on this, an advection equation is introduced to progressively refine both the feature space and the semantic response space. Finally, entity-level FSS is designed to produce the final segmentation results.

\subsection{Category-Agnostic Entity Primitive Generation}
\label{SEC_2_2}
Remote sensing objects generally exhibit strong spatial continuity, with pixels belonging to the same object sharing highly consistent semantics. Conventional pixel-wise prediction is easily affected by local texture variations and spectral fluctuations, often resulting in local holes and fragmented misclassifications within homogeneous regions. In contrast, aggregating semantic information over spatially continuous regions effectively alleviates these issues. Motivated by this observation, the proposed method exploits the generic region modeling capability of SAM3 to partition the query image into a set of category-agnostic entity primitives. These entity primitives replace individual pixels as the basic inference units, thereby transforming FSS from dense pixel-wise prediction into entity-level inference.

Specifically, the automatic mask generation module of SAM3 first produces a set of candidate masks together with their corresponding quality scores. Candidate regions with excessively small areas are discarded, and the remaining masks are ranked in descending order according to their quality scores. To resolve the widespread overlap and inclusion relationships among candidate masks, a confidence-priority greedy exclusive assignment strategy is adopted. Pixels that have already been assigned to higher-confidence entities are progressively removed from subsequent candidate masks, and only the remaining valid regions satisfying the minimum area constraint are retained as entity primitives. After all candidate masks have been processed, the remaining unassigned pixels are merged into a residual entity primitive. Consequently, the resulting entity primitives are spatially disjoint while jointly covering the entire query image. Entity primitives generated from candidate masks inherit the corresponding SAM3 quality scores, whereas the residual entity primitive is assigned a quality score of zero. These category-agnostic entity primitives provide only spatial and geometric priors, while their semantic categories are determined during the subsequent entity-level semantic inference stage. The details of category-agnostic entity primitive generation are presented in \textit{Appendix}.

\subsection{Multimodal Semantic Potential Field Construction}
\label{sec:potential}
After generating the category-agnostic entity primitives, we construct a category-aware continuous semantic potential field for each target category by integrating target-domain support samples and category text prompts. The semantic potential field serves as the guidance signal for the subsequent semantic advection refinement.

For category $c$, the $k$-th support image $I_s^{(c,k)}$ is first encoded by the frozen SAM3 backbone to obtain the visual feature map $F_s^{(c,k)}\in\mathbb{R}^{D\times H\times W}$, where $D$ denotes the feature dimension. Given the corresponding binary support mask $M_s^{(c,k)}$, masked average pooling followed by $L_2$ normalization is performed over the foreground region of each support image. The foreground prototype of category $c$ is then computed as
\begin{equation}
\small
P_c^{\mathrm{fg}}=\operatorname{Norm}\left[\frac{1}{K}\sum_{k=1}^{K}\operatorname{Norm}
\left(\frac{\sum_{px\in\Omega_s}M_s^{(c,k)}(px)F_s^{(c,k)}(px)}{\sum_{px\in\Omega_s}M_s^{(c,k)}(px)+\varepsilon}
\right)\right],
\label{eq:fg_proto}
\end{equation}
where $\Omega_s$ denotes the pixel domain of the support image, $\varepsilon$ is a small constant introduced for numerical stability, and $\operatorname{Norm}(\cdot)$ denotes $L_2$ normalization. Similarly, replacing the foreground mask $M_s^{(c,k)}$ with its complementary background mask $1-M_s^{(c,k)}$ yields the background prototype $P_c^{\mathrm{bg}}$. The foreground prototype represents the characteristic visual appearance of the target category, whereas the background prototype captures the visual characteristics of non-target regions relative to that category.

Then, the query image $I_q$ together with the category prompt $t_c$ is fed into SAM3 to obtain the category-conditioned query feature map $F_q^c\in\mathbb{R}^{D\times H\times W}$, and the dense text semantic response map $R_c^{\mathrm{txt},(0)}$. For each query pixel $px\in\Omega_q$, the cosine similarities between the query feature $F_q^c(px)$ and the foreground and background prototypes are respectively computed as
\begin{equation}
R_c^{\mathrm{fg},(0)}(px)=\frac{{F_q^c(px)}^{\!\top}P_c^{\mathrm{fg}}}{\|F_q^c(px)\|_2\, \|P_c^{\mathrm{fg}}\|_2},
\label{eq:fg_response}
\end{equation}
and
\begin{equation}
R_c^{\mathrm{bg},(0)}(px)=\frac{{F_q^c(px)}^{\!\top}P_c^{\mathrm{bg}}}{\|F_q^c(px)\|_2\,\|P_c^{\mathrm{bg}}\|_2}.
\label{eq:bg_response}
\end{equation}

Based on the foreground, background, and textual semantic responses, a feature-level semantic potential field for category $c$ is constructed. Specifically, the visual confidence that each query pixel $px$ belongs to the foreground of category $c$ is first obtained through Softmax normalization:
\begin{equation}
p_{c}^{\mathrm{vis},F}(px)=\frac{\exp\left(R_{c}^{\mathrm{fg},(0)}(px)\right)}{\exp\left(R_{c}^{\mathrm{fg},(0)}(px)\right)+\exp\left(R_{c}^{\mathrm{bg},(0)}(px)\right)}.
\label{eq:visual_confidence}
\end{equation}

The dense text-semantic response map $R_{c}^{\mathrm{txt},(0)}$ is then employed as a position-dependent semantic modulation factor to adaptively adjust the visual foreground confidence. Accordingly, the feature-level semantic potential field for category $c$ is constructed as
\begin{equation}
U_{c}^{F}(px)=p_{c}^{\mathrm{vis},F}(px)\left[(1-\alpha^{F})+\alpha^{F}R_{c}^{\mathrm{txt},(0)}(px)\right],
\label{eq:feature_semantic_potential}
\end{equation}
where $\alpha^{F}$ denotes the modulation strength of the feature-level textual semantics and is set to $0.25$ in our experiments. The resulting semantic potential field
$U_{c}^{F}\in[0,1]^{H\times W}$
jointly encodes the foreground-background visual discrimination information of category $c$ and its category-related textual semantic information, and is subsequently used to construct the category-specific semantic velocity field.

\subsection{Semantic Advection Refinement}
\label{sec:advection}
After constructing the category-aware multimodal semantic potential field, we introduce a two-dimensional advection equation to perform lightweight spatial advection on the query features and multi-source semantic responses. The proposed refinement propagates category-related information in a directionally controlled manner by exploiting the spatial gradients of the semantic potential field.

\subsubsection{Query Feature Advection Refinement}
\label{sec:advection_1}
Let $D_x$ and $D_y$ denote the difference operators along the horizontal and vertical directions, respectively. For a two-dimensional field $Z$, the horizontal difference operator is defined as
\begin{equation}
D_x Z(x,y)=
\begin{cases}
Z(x+1,y)-Z(x,y), & x=1,\\[2mm]
\dfrac{Z(x+1,y)-Z(x-1,y)}{2}, & 1<x<W_q,\\[3mm]
Z(x,y)-Z(x-1,y), & x=W_q,
\end{cases}
\label{eq:dx}
\end{equation}
where $W_q$ denotes the width of the query image. The vertical operator $D_y$ applies the same finite difference scheme as $D_x$, but along the orthogonal direction. 

Based on the feature-level semantic potential field $U_c^{F}$, we construct the normalized semantic velocity field $\mathbf{v}_c^{F}=\left(v_{c,x}^{F}, v_{c,y}^{F}\right)$, where
\begin{equation}
v_{c,x}^{F}=-\frac{D_xU_c^{F}(x,y)}{\sqrt{(D_xU_c^{F})^2+(D_yU_c^{F})^2}+\varepsilon},
\label{eq:vx}
\end{equation}
and
\begin{equation}
v_{c,y}^{F}=-\frac{D_yU_c^{F}(x,y)}{\sqrt{(D_xU_c^{F})^2+(D_yU_c^{F})^2}+\varepsilon}.
\label{eq:vy}
\end{equation}

Let $F_q^{c,(t)}$ denote the query feature after the $t$-th iteration, with the initialization $F_q^{c,(0)}=F_q^{c}$. The feature evolution is assumed to satisfy the following two-dimensional advection equation:
\begin{equation}
\frac{\partial F_q^c}{\partial t}+v_{c,x}^{F}\frac{\partial F_q^c}{\partial x}+v_{c,y}^{F}\frac{\partial F_q^c}{\partial y}=0.
\label{eq:feature_pde}
\end{equation}

At the $t$-th iteration, the feature-level semantic potential field $U_{c}^{F,(t)}$ is recomputed from the current query feature $F_q^{c,(t)}$. The directional derivative of the semantic potential along the fixed velocity field is computed as
\begin{equation}
d_{c}^{F,(t)}=v_{c,x}^{F}D_xU_{c}^{F,(t)}+v_{c,y}^{F}D_yU_{c}^{F,(t)}. 
\label{eq:feature_directional}
\end{equation}
The semantic directional gate is then defined as
\begin{equation}
G_{c}^{F,(t)}=\mathbf{1}\!\left(d_{c}^{F,(t)}<0\right),
\label{eq:feature_gate}
\end{equation}
where $\mathbf{1}(\cdot)$ denotes the indicator function, which equals $1$ if $d_{c}^{F,(t)}<0$, and $0$ otherwise. The gate is activated only when the current transport direction enhances the semantic response of class $c$, thereby suppressing advection updates that deviate from the desired semantic direction. Then, the above equation is numerically solved using the explicit Euler scheme. The intermediate feature obtained at the $(t+1)$-th iteration is computed as
\begin{equation}
\tilde{F}_q^{c,(t+1)}=F_q^{c,(t)}-\delta^F G_{c}^{F,(t)}\left(v_{c,x}^{F}D_x F_q^{c,(t)}+v_{c,y}^{F}D_y F_q^{c,(t)}\right),
\label{eq:feature_euler}
\end{equation}
where $\delta^F$ denotes the feature-advection time step, which is empirically set to 0.06 throughout the experiments.

To limit the update magnitude while preserving the original feature structure, a residual interpolation strategy is adopted:
\begin{equation}
\bar{F}_q^{c,(t+1)} = \left[(1-\gamma^F)F_q^{c,(t)}+\gamma^F\tilde{F}_q^{c,(t+1)}\right], 
\label{eq:feature_update}
\end{equation}
where $\gamma^F$ controls the feature-advection strength. When the feature on a position $px$ is not changed because of the gate (\ref{eq:feature_gate}), we have $\tilde{F}_q^{c,(t+1)}(px) = \bar{F}_q^{c,(t+1)}(px) = F_q^{c,(t)}(px)$ based on (\ref{eq:feature_update}). In all experiments, we set $\gamma^F=0.18$. After $N^F$ iterations, the refined query feature is obtained and normalized using channel-wise $L_2$ normalization as $\hat{F}_q^{\,c}=\operatorname{Norm}(\bar{F}_q^{c,(N^F)})$, where the number of feature-advection iterations is empirically fixed to $N^F=2$.

\subsubsection{Multi-Source Response Advection Refinement}
\label{sec:advection_2}
Although the first-stage advection has propagated category-related information within the high-dimensional visual feature space, the recomputed foreground, background, and textual responses may still contain isolated activations or local discontinuities caused by imperfect local matching and spatial sampling artifacts. Therefore, the second-stage refinement which operates directly on the semantic response maps to further improve the spatial continuity is designed.

After completing query feature advection refinement, the cosine similarities between the refined query feature $\hat{F}_q^{\,c}$ and the foreground and background prototypes are recomputed following Eqs.~(\ref{eq:fg_response})--(\ref{eq:bg_response}), yielding the refined foreground and background response maps, denoted by $\hat{R}_c^{\mathrm{fg}}(x)$ and $\hat{R}_c^{\mathrm{bg}}(x)$, respectively.

For each response type $u\in\{\mathrm{fg},\mathrm{bg},\mathrm{txt}\}$, the feature-refined response is used as the initial condition of the response-level advection process, namely, $R_c^{u,(0)}=\hat{R}_c^{u}$. Based on these responses, the response-level semantic potential field $U_c^{R}(x)$ is constructed following the same formulation as Eqs.~(\ref{eq:visual_confidence})--(\ref{eq:feature_semantic_potential}). Similarly, using the response-level semantic potential field $U_c^{R}$, the normalized response-level semantic velocity field $\mathbf{v}_c^{R}=\left(v_{c,x}^{R},v_{c,y}^{R}\right)$ is computed in the same manner as Eqs.~(\ref{eq:vx})--(\ref{eq:vy}).

Analogous to Eq.~(\ref{eq:feature_pde}), the response evolution is modeled by the following two-dimensional advection equation:
\begin{equation}
\frac{\partial R_c^{u}}{\partial t}+v_{c,x}^{R}\frac{\partial R_c^{u}}{\partial x}+v_{c,y}^{R}\frac{\partial R_c^{u}}{\partial y}=0,
\ \ u\in\{\mathrm{fg},\mathrm{bg},\mathrm{txt}\}.
\label{eq:response_pde}
\end{equation}

We further introduce response-specific semantic directional gating similar to Eqs.~(\ref{eq:feature_directional}-\ref{eq:feature_gate}). Then, the above equation is discretized using the same explicit Euler scheme as Eq.~(\ref{eq:feature_euler}). Let $\tilde{R}_c^{u,(t+1)}$ denote the intermediate response obtained at the $(t+1)$-th iteration. To limit the update magnitude while maintaining the numerical stability of the response range, residual interpolation followed by value clipping is performed:
{\small
\begin{equation}
R_c^{u,(t+1)}=\min\left\{1,\max\left[-1,(1-\gamma^R)R_c^{u,(t)}+ \gamma^R\tilde{R}_c^{u,(t+1)}\right]\right\}.
\label{eq:response_update}
\end{equation}
}
Here, $\gamma^R$ controls the response-advection strength and is empirically set to $0.18$. After $N^R=2$ iterations, the original response and the advection-refined response are linearly fused as
\begin{equation}
\bar{R}_c^{u}=\rho\hat{R}_c^{u}+(1-\rho)R_c^{u,(N^R)},
\qquad u\in\{\mathrm{fg},\mathrm{bg},\mathrm{txt}\},
\label{eq:response_fusion}
\end{equation}
where $\rho$ denotes the retention weight of the original response and is empirically set to $0.50$.

\subsection{Entity-Level Semantic Inference}
\label{sec:inference}
After semantic advection refinement, category-agnostic entity primitives are treated as the minimum segmentation units for entity-level inference. For each entity primitive $g_i$ and target category $c$, the refined foreground, background, and textual semantic responses are first aggregated within the entity region as
\begin{equation}
\mu_{i,c}^{u}=\frac{\sum_{x\in\Omega_q}\bar{R}_c^{u}(x)\,g_i(x)}{\sum_{x\in\Omega_q}g_i(x)+\varepsilon},
\ \ u\in\{\mathrm{fg},\mathrm{bg},\mathrm{txt}\},
\label{eq:entity_response}
\end{equation}
where $\Omega_q$ denotes the pixel domain of the query image.

The average response-level semantic potential within the entity primitive is defined as
\begin{equation}
\mu_{i,c}^{\mathrm{sem}}=\frac{\sum_{x\in\Omega_q}U_c^{R}(x)\,g_i(x)}{\sum_{x\in\Omega_q}g_i(x)+\varepsilon}.
\label{eq:entity_sem}
\end{equation}

Based on the above intermediate quantities, the fused semantic score of entity primitive $g_i$ with respect to category $c$ is computed as
\begin{equation}
\begin{aligned}
z_{i,c}={}& w_{\mathrm{fg}}\frac{\mu_{i,c}^{\mathrm{fg}}+1}{2}-w_{\mathrm{bg}}\frac{\mu_{i,c}^{\mathrm{bg}}+1}{2}+w_{\mathrm{txt}}\frac{\mu_{i,c}^{\mathrm{txt}}+1}{2}
\\
&+w_{\mathrm{sem}}\mu_{i,c}^{\mathrm{sem}}+w_{\mathrm{sam}}s_i^{\mathrm{sam}},
\end{aligned}
\label{eq:fusion_score}
\end{equation}
where $s_i^{\mathrm{sam}}$ is a quality score which is described in the “Details of Category-Agnostic Entity Primitive Generation” section of the \textit{Appendix}. Here, $w_{\mathrm{fg}}=1.25$, $w_{\mathrm{bg}}=1.35$, $w_{\mathrm{txt}}=0.03$, $w_{\mathrm{sem}}=0.25$ and $w_{\mathrm{sam}}=0.01$.

The category confidence $p_{i,c}$ is then computed using a sigmoid function taking $z_{i,c}$ as input. For each entity primitive, the category with the highest confidence score is selected as its predicted label:
\begin{equation}
c_i^{*}=\arg\max_{c\in\{1,\ldots,C\}}p_{i,c}.
\label{eq:prediction}
\end{equation}

Finally, the predicted category label is assigned to all pixels covered by the corresponding entity primitive, producing the final result of the query image.

\section{Experiment}
\label{sec:exp}
To comprehensively evaluate the effectiveness of the proposed method across different sensors, spatial resolutions, and complex land-cover/land-use scenarios, extensive experiments are conducted on seven representative and highly challenging remote sensing datasets, including GID-15 \cite{TONG-2020}, Five-Billion-Pixels (FBP) \cite{TONG-2023}, Potsdam \cite{ISPRSDATA-2014}, Vaihingen \cite{ISPRSDATA-2014}, iSAID \cite{iSAID-2019}, FLAIR \#1 \cite{FLAIR-2022}, and MiniFrance \cite{MiniFrance-2020}. For all datasets, the official category names are directly adopted as the textual prompts, the mean Intersection over Union (mIoU) and Boundary Intersection over Union (BIoU) are adopted as the evaluation metrics. The detailed experimental settings are presented in \textit{Appendix}. 

\subsection{Comparisons}
\label{sec:exp_comp}

\begin{table*}[t]
\centering
\caption{Cross-domain few-shot segmentation results under training on the PASCAL~\cite{Pascal-2015} source domain. The best results are highlighted in bold, and standard deviations are shown in blue.}
\label{tab:pascal_cdfss}
\resizebox{\textwidth}{!}{
\begin{tabular}{llcccccccccccccccc}
\toprule
\multirow{2}{*}{Method} & \multirow{2}{*}{Category} &
\multicolumn{2}{c}{Potsdam} & \multicolumn{2}{c}{Vaihingen} &
\multicolumn{2}{c}{iSAID} & \multicolumn{2}{c}{GID-15} &
\multicolumn{2}{c}{FBP} & \multicolumn{2}{c}{FLAIR \#1} &
\multicolumn{2}{c}{MiniFrance} & \multicolumn{2}{c}{Average} \\
\cmidrule(lr){3-4}\cmidrule(lr){5-6}\cmidrule(lr){7-8}
\cmidrule(lr){9-10}\cmidrule(lr){11-12}\cmidrule(lr){13-14}
\cmidrule(lr){15-16}\cmidrule(lr){17-18}
& & 1-shot & 5-shot & 1-shot & 5-shot & 1-shot & 5-shot &
1-shot & 5-shot & 1-shot & 5-shot & 1-shot & 5-shot &
1-shot & 5-shot & 1-shot & 5-shot \\
\midrule
PANet~\cite{PANet-2019} & \multirow{5}{*}{FSS} &
\makecell{37.08 \\ \textcolor{blue}{\(\pm\) 2.51}} &
\makecell{42.40 \\ \textcolor{blue}{\(\pm\) 2.28}} &
\makecell{35.69 \\ \textcolor{blue}{\(\pm\) 2.67}} &
\makecell{40.85 \\ \textcolor{blue}{\(\pm\) 2.23}} &
\makecell{13.63 \\ \textcolor{blue}{\(\pm\) 2.78}} &
\makecell{13.65 \\ \textcolor{blue}{\(\pm\) 2.76}} &
\makecell{11.34 \\ \textcolor{blue}{\(\pm\) 2.64}} &
\makecell{14.05 \\ \textcolor{blue}{\(\pm\) 2.56}} &
\makecell{7.19 \\ \textcolor{blue}{\(\pm\) 2.76}} &
\makecell{10.63 \\ \textcolor{blue}{\(\pm\) 2.32}} &
\makecell{13.83 \\ \textcolor{blue}{\(\pm\) 2.75}} &
\makecell{16.72 \\ \textcolor{blue}{\(\pm\) 2.15}} &
\makecell{8.47 \\ \textcolor{blue}{\(\pm\) 2.47}} &
\makecell{9.02 \\ \textcolor{blue}{\(\pm\) 2.08}} &
18.18 & 21.05 \\

PFENet~\cite{PFENet-2022} & &
\makecell{23.96 \\ \textcolor{blue}{\(\pm\) 2.63}} &
\makecell{24.20 \\ \textcolor{blue}{\(\pm\) 2.33}} &
\makecell{19.03 \\ \textcolor{blue}{\(\pm\) 2.71}} &
\makecell{20.38 \\ \textcolor{blue}{\(\pm\) 2.18}} &
\makecell{9.09 \\ \textcolor{blue}{\(\pm\) 2.94}} &
\makecell{9.92 \\ \textcolor{blue}{\(\pm\) 2.80}} &
\makecell{8.42 \\ \textcolor{blue}{\(\pm\) 2.63}} &
\makecell{10.26 \\ \textcolor{blue}{\(\pm\) 2.59}} &
\makecell{6.65 \\ \textcolor{blue}{\(\pm\) 2.84}} &
\makecell{8.72 \\ \textcolor{blue}{\(\pm\) 2.30}} &
\makecell{12.92 \\ \textcolor{blue}{\(\pm\) 2.89}} &
\makecell{14.54 \\ \textcolor{blue}{\(\pm\) 2.20}} &
\makecell{7.21 \\ \textcolor{blue}{\(\pm\) 2.61}} &
\makecell{8.20 \\ \textcolor{blue}{\(\pm\) 1.95}} &
12.47 & 13.75 \\

SSP~\cite{SSP-2022} & &
\makecell{36.34 \\ \textcolor{blue}{\(\pm\) 2.71}} &
\makecell{42.68 \\ \textcolor{blue}{\(\pm\) 2.30}} &
\makecell{37.58 \\ \textcolor{blue}{\(\pm\) 2.64}} &
\makecell{40.65 \\ \textcolor{blue}{\(\pm\) 2.21}} &
\makecell{14.75 \\ \textcolor{blue}{\(\pm\) 2.73}} &
\makecell{17.22 \\ \textcolor{blue}{\(\pm\) 2.64}} &
\makecell{11.58 \\ \textcolor{blue}{\(\pm\) 2.60}} &
\makecell{13.45 \\ \textcolor{blue}{\(\pm\) 2.55}} &
\makecell{6.51 \\ \textcolor{blue}{\(\pm\) 2.75}} &
\makecell{9.17 \\ \textcolor{blue}{\(\pm\) 2.29}} &
\makecell{13.36 \\ \textcolor{blue}{\(\pm\) 2.68}} &
\makecell{16.05 \\ \textcolor{blue}{\(\pm\) 2.10}} &
\makecell{8.80 \\ \textcolor{blue}{\(\pm\) 2.49}} &
\makecell{9.46 \\ \textcolor{blue}{\(\pm\) 1.93}} &
18.42 & 21.24 \\
\midrule
IFA~\cite{CDFSS-Nie2024} & \multirow{7}{*}{CDFSS} &
\makecell{39.52 \\ \textcolor{blue}{\(\pm\) 2.49}} &
\makecell{45.96 \\ \textcolor{blue}{\(\pm\) 2.20}} &
\makecell{40.92 \\ \textcolor{blue}{\(\pm\) 2.58}} &
\makecell{42.67 \\ \textcolor{blue}{\(\pm\) 2.15}} &
\makecell{16.02 \\ \textcolor{blue}{\(\pm\) 2.78}} &
\makecell{19.68 \\ \textcolor{blue}{\(\pm\) 2.62}} &
\makecell{14.79 \\ \textcolor{blue}{\(\pm\) 2.61}} &
\makecell{17.01 \\ \textcolor{blue}{\(\pm\) 2.52}} &
\makecell{7.95 \\ \textcolor{blue}{\(\pm\) 2.47}} &
\makecell{10.90 \\ \textcolor{blue}{\(\pm\) 2.25}} &
\makecell{16.24 \\ \textcolor{blue}{\(\pm\) 2.44}} &
\makecell{19.38 \\ \textcolor{blue}{\(\pm\) 2.07}} &
\makecell{10.52 \\ \textcolor{blue}{\(\pm\) 2.43}} &
\makecell{11.27 \\ \textcolor{blue}{\(\pm\) 1.92}} &
20.85 & 23.84 \\

DR-Adapter~\cite{CDFSS-Su2024} & &
\makecell{37.16 \\ \textcolor{blue}{\(\pm\) 2.57}} &
\makecell{44.63 \\ \textcolor{blue}{\(\pm\) 2.25}} &
\makecell{38.16 \\ \textcolor{blue}{\(\pm\) 2.51}} &
\makecell{40.37 \\ \textcolor{blue}{\(\pm\) 2.13}} &
\makecell{15.72 \\ \textcolor{blue}{\(\pm\) 2.62}} &
\makecell{17.49 \\ \textcolor{blue}{\(\pm\) 2.47}} &
\makecell{12.59 \\ \textcolor{blue}{\(\pm\) 2.66}} &
\makecell{16.49 \\ \textcolor{blue}{\(\pm\) 2.53}} &
\makecell{9.30 \\ \textcolor{blue}{\(\pm\) 2.53}} &
\makecell{11.36 \\ \textcolor{blue}{\(\pm\) 2.24}} &
\makecell{15.74 \\ \textcolor{blue}{\(\pm\) 2.51}} &
\makecell{20.05 \\ \textcolor{blue}{\(\pm\) 2.12}} &
\makecell{8.95 \\ \textcolor{blue}{\(\pm\) 2.38}} &
\makecell{11.74 \\ \textcolor{blue}{\(\pm\) 2.00}} &
19.66 & 23.16 \\

APM~\cite{APM-2024} & &
\makecell{36.59 \\ \textcolor{blue}{\(\pm\) 2.53}} &
\makecell{42.11 \\ \textcolor{blue}{\(\pm\) 2.27}} &
\makecell{36.69 \\ \textcolor{blue}{\(\pm\) 2.62}} &
\makecell{42.13 \\ \textcolor{blue}{\(\pm\) 2.11}} &
\makecell{14.70 \\ \textcolor{blue}{\(\pm\) 2.49}} &
\makecell{17.84 \\ \textcolor{blue}{\(\pm\) 2.38}} &
\makecell{12.27 \\ \textcolor{blue}{\(\pm\) 2.82}} &
\makecell{15.01 \\ \textcolor{blue}{\(\pm\) 2.56}} &
\makecell{7.98 \\ \textcolor{blue}{\(\pm\) 2.68}} &
\makecell{10.81 \\ \textcolor{blue}{\(\pm\) 2.28}} &
\makecell{14.40 \\ \textcolor{blue}{\(\pm\) 2.62}} &
\makecell{18.82 \\ \textcolor{blue}{\(\pm\) 2.19}} &
\makecell{8.24 \\ \textcolor{blue}{\(\pm\) 2.56}} &
\makecell{10.75 \\ \textcolor{blue}{\(\pm\) 2.03}} &
18.70 & 22.50 \\

ICPD~\cite{ICPD-2025} & &
\makecell{33.42 \\ \textcolor{blue}{\(\pm\) 2.65}} &
\makecell{36.73 \\ \textcolor{blue}{\(\pm\) 2.21}} &
\makecell{36.11 \\ \textcolor{blue}{\(\pm\) 2.71}} &
\makecell{38.56 \\ \textcolor{blue}{\(\pm\) 2.18}} &
\makecell{15.30 \\ \textcolor{blue}{\(\pm\) 2.77}} &
\makecell{16.10 \\ \textcolor{blue}{\(\pm\) 2.61}} &
\makecell{12.68 \\ \textcolor{blue}{\(\pm\) 2.70}} &
\makecell{14.58 \\ \textcolor{blue}{\(\pm\) 2.55}} &
\makecell{7.26 \\ \textcolor{blue}{\(\pm\) 2.54}} &
\makecell{9.36 \\ \textcolor{blue}{\(\pm\) 2.30}} &
\makecell{14.65 \\ \textcolor{blue}{\(\pm\) 2.73}} &
\makecell{16.91 \\ \textcolor{blue}{\(\pm\) 2.14}} &
\makecell{9.28 \\ \textcolor{blue}{\(\pm\) 2.49}} &
\makecell{11.54 \\ \textcolor{blue}{\(\pm\) 1.90}} &
18.39 & 20.54 \\
\midrule
FSSAM~\cite{Xu-2025} & Foundation-based FSS &
\makecell{23.53 \\ \textcolor{blue}{\(\pm\) 2.57}} &
\makecell{24.13 \\ \textcolor{blue}{\(\pm\) 2.24}} &
\makecell{27.86 \\ \textcolor{blue}{\(\pm\) 2.43}} &
\makecell{31.65 \\ \textcolor{blue}{\(\pm\) 2.08}} &
\makecell{7.89 \\ \textcolor{blue}{\(\pm\) 2.68}} &
\makecell{7.68 \\ \textcolor{blue}{\(\pm\) 2.28}} &
\makecell{8.87 \\ \textcolor{blue}{\(\pm\) 2.65}} &
\makecell{10.94 \\ \textcolor{blue}{\(\pm\) 2.58}} &
\makecell{5.13 \\ \textcolor{blue}{\(\pm\) 2.74}} &
\makecell{7.23 \\ \textcolor{blue}{\(\pm\) 2.28}} &
\makecell{14.74 \\ \textcolor{blue}{\(\pm\) 2.48}} &
\makecell{17.28 \\ \textcolor{blue}{\(\pm\) 2.08}} &
\makecell{9.35 \\ \textcolor{blue}{\(\pm\) 2.43}} &
\makecell{11.91 \\ \textcolor{blue}{\(\pm\) 1.94}} &
13.91 & 15.83 \\
\midrule
\textbf{ELFSS-AR (Ours)} & Training-free CDFSS &
\makecell{\textbf{54.27} \\ \textcolor{blue}{\(\pm\) 2.33}} &
\makecell{\textbf{60.61} \\ \textcolor{blue}{\(\pm\) 2.15}} &
\makecell{\textbf{57.39} \\ \textcolor{blue}{\(\pm\) 2.40}} &
\makecell{\textbf{63.37} \\ \textcolor{blue}{\(\pm\) 2.05}} &
\makecell{\textbf{21.04} \\ \textcolor{blue}{\(\pm\) 2.45}} &
\makecell{\textbf{22.58} \\ \textcolor{blue}{\(\pm\) 2.18}} &
\makecell{\textbf{19.56} \\ \textcolor{blue}{\(\pm\) 2.60}} &
\makecell{\textbf{26.16} \\ \textcolor{blue}{\(\pm\) 2.49}} &
\makecell{\textbf{15.35} \\ \textcolor{blue}{\(\pm\) 2.47}} &
\makecell{\textbf{17.67} \\ \textcolor{blue}{\(\pm\) 2.24}} &
\makecell{\textbf{18.97} \\ \textcolor{blue}{\(\pm\) 2.36}} &
\makecell{\textbf{23.78} \\ \textcolor{blue}{\(\pm\) 2.01}} &
\makecell{\textbf{12.24} \\ \textcolor{blue}{\(\pm\) 2.05}} &
\makecell{\textbf{14.81} \\ \textcolor{blue}{\(\pm\) 1.87}} &
\textbf{28.40} & \textbf{32.71} \\
\bottomrule
\end{tabular}}
\end{table*}

\begin{table*}[t]
\centering
\caption{Comparison with open-vocabulary semantic segmentation methods based on frozen foundation models. Except for ConceptBank, the remaining training-free OVSS methods use only the official category names as textual prompts; ConceptBank further leverages target-domain support images to construct a concept bank. With all SAM3 parameters frozen, our method additionally uses one or five target-domain support images and their corresponding masks for each category. The best results are highlighted in bold, and standard deviations are shown in blue.}
\label{tab:training_free_ovss}
\resizebox{0.92\textwidth}{!}{
\begin{tabular}{llcccccccc}
\toprule
Method & Setting & Potsdam & Vaihingen & iSAID & GID-15 &
FBP & FLAIR \#1 & MiniFrance & Average \\
\midrule
MaskCLIP~\cite{Zhou-2022} & Training-free OVSS &
26.54 & 24.80 & 5.61 & 18.73 & 14.07 & 17.22 & 11.68 & 16.95 \\
ClearCLIP~\cite{ClearCLIP-2025} & Training-free OVSS &
26.29 & 22.01 & 5.53 & 19.52 & 13.56 & 17.89 & 11.15 & 16.56 \\
SCLIP~\cite{SCLIP-2025} & Training-free OVSS &
27.13 & 22.72 & 5.25 & 19.45 & 13.33 & 16.93 & 11.84 & 16.66 \\
ResCLIP~\cite{ResCLIP-2025} & Training-free OVSS &
27.02 & 23.39 & 5.09 & 18.95 & 14.79 & 18.05 & 12.01 & 17.04 \\
SegEarth-OV~\cite{SegEarth-OV-2025} & Training-free OVSS &
30.63 & 23.03 & 14.82 & 9.73 & 10.99 & 18.48 & 10.46 & 16.88 \\
SAM3~\cite{SAM3-2026} & Training-free OVSS &
49.19 & 38.28 & 6.37 & 8.33 & 5.30 & 14.75 & 7.72 & 18.56 \\
\midrule
\multirow{3}{*}{ConceptBank~\cite{conceptbank_2026}} & Support-assisted OVSS (1-shot) &
\makecell{{45.66} \\ \textcolor{blue}{\(\pm\) 2.40}} & \makecell{{40.81} \\ \textcolor{blue}{\(\pm\) 2.46}} & \makecell{{13.39} \\ \textcolor{blue}{\(\pm\) 2.64}} & \makecell{{7.86} \\ \textcolor{blue}{\(\pm\) 2.52}} & \makecell{{10.24} \\ \textcolor{blue}{\(\pm\) 2.26}} & \makecell{{16.75} \\ \textcolor{blue}{\(\pm\) 2.35}} & \makecell{{7.33} \\ \textcolor{blue}{\(\pm\) 1.96}} & 20.29 \\
& Support-assisted OVSS (5-shot) &
\makecell{{51.06} \\ \textcolor{blue}{\(\pm\) 2.17}} & \makecell{{45.62} \\ \textcolor{blue}{\(\pm\) 2.32}} & \makecell{{15.70} \\ \textcolor{blue}{\(\pm\) 2.57}} & \makecell{{9.64} \\ \textcolor{blue}{\(\pm\) 2.36}} & \makecell{{11.63} \\ \textcolor{blue}{\(\pm\) 2.26}} & \makecell{{19.16} \\ \textcolor{blue}{\(\pm\) 2.06}} & \makecell{{8.59} \\ \textcolor{blue}{\(\pm\) 1.92}} & 23.06 \\
\midrule
\multirow{3}{*}{\textbf{ELFSS-AR (Ours)}} & 1-shot &
\makecell{\textbf{54.27} \\ \textcolor{blue}{\(\pm\) 2.33}} & \makecell{\textbf{57.39} \\ \textcolor{blue}{\(\pm\) 2.40}} & \makecell{\textbf{21.04} \\ \textcolor{blue}{\(\pm\) 2.45}} &
\makecell{\textbf{19.56} \\ \textcolor{blue}{\(\pm\) 2.60}} &\makecell{\textbf{15.35} \\ \textcolor{blue}{\(\pm\) 2.47}} &\makecell{\textbf{18.97} \\ \textcolor{blue}{\(\pm\) 2.36}} &\makecell{\textbf{12.24} \\ \textcolor{blue}{\(\pm\) 2.05}} & \textbf{28.40} \\
& 5-shot &
\makecell{\textbf{60.61} \\ \textcolor{blue}{\(\pm\) 2.15}} & \makecell{\textbf{63.37} \\ \textcolor{blue}{\(\pm\) 2.05}} & \makecell{\textbf{22.58} \\ \textcolor{blue}{\(\pm\) 2.18}} &
\makecell{\textbf{26.16} \\ \textcolor{blue}{\(\pm\) 2.49}} & \makecell{\textbf{17.67} \\ \textcolor{blue}{\(\pm\) 2.24}} & \makecell{\textbf{23.78} \\ \textcolor{blue}{\(\pm\) 2.01}} & \makecell{\textbf{14.81} \\ \textcolor{blue}{\(\pm\) 1.87}} & \textbf{32.71} \\
\bottomrule
\end{tabular}}
\end{table*}

\begin{figure}[t]
\centering
\includegraphics[width=0.47\textwidth]{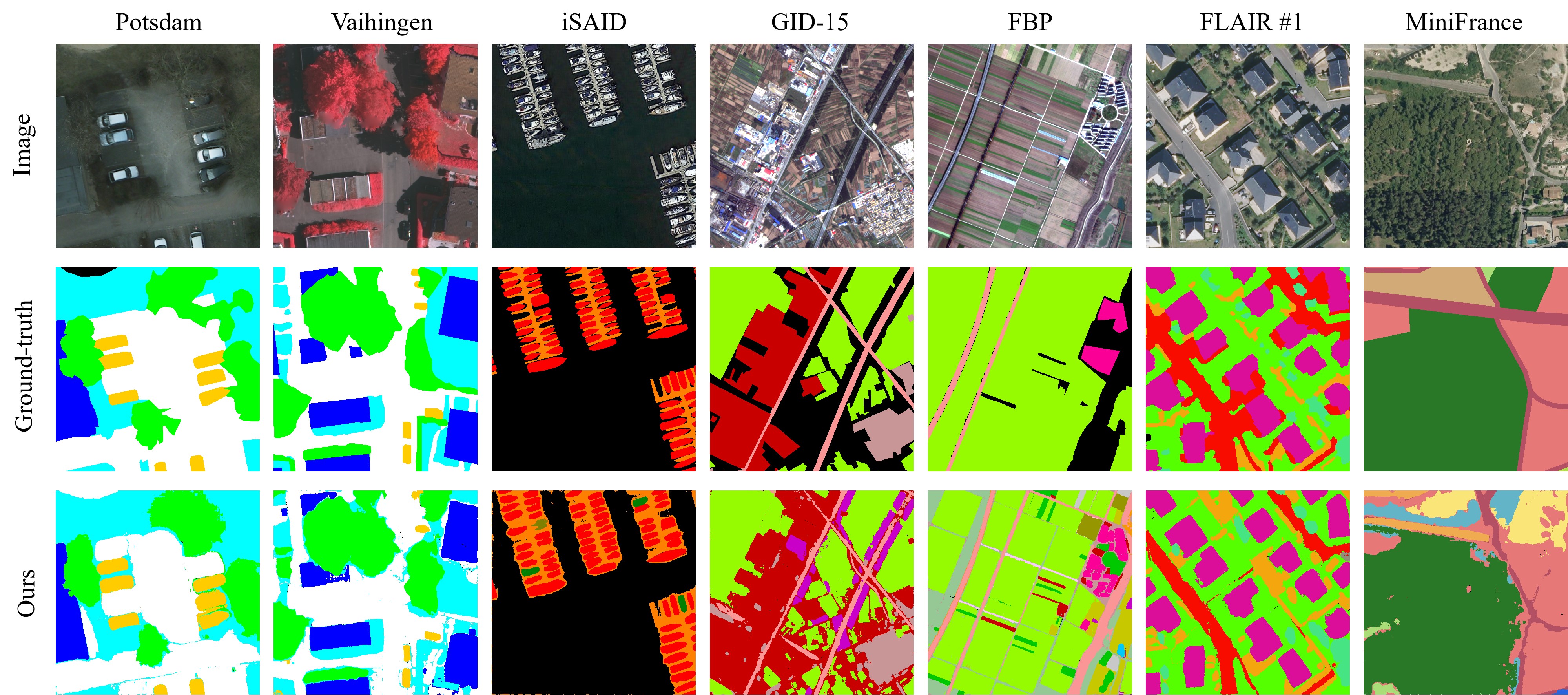}
\caption{Segmentation results under the 5-shot setting. }
\label{fig:qualitative_results}
\end{figure}

\begin{figure*}[t]
\centering
\includegraphics[width=\textwidth]{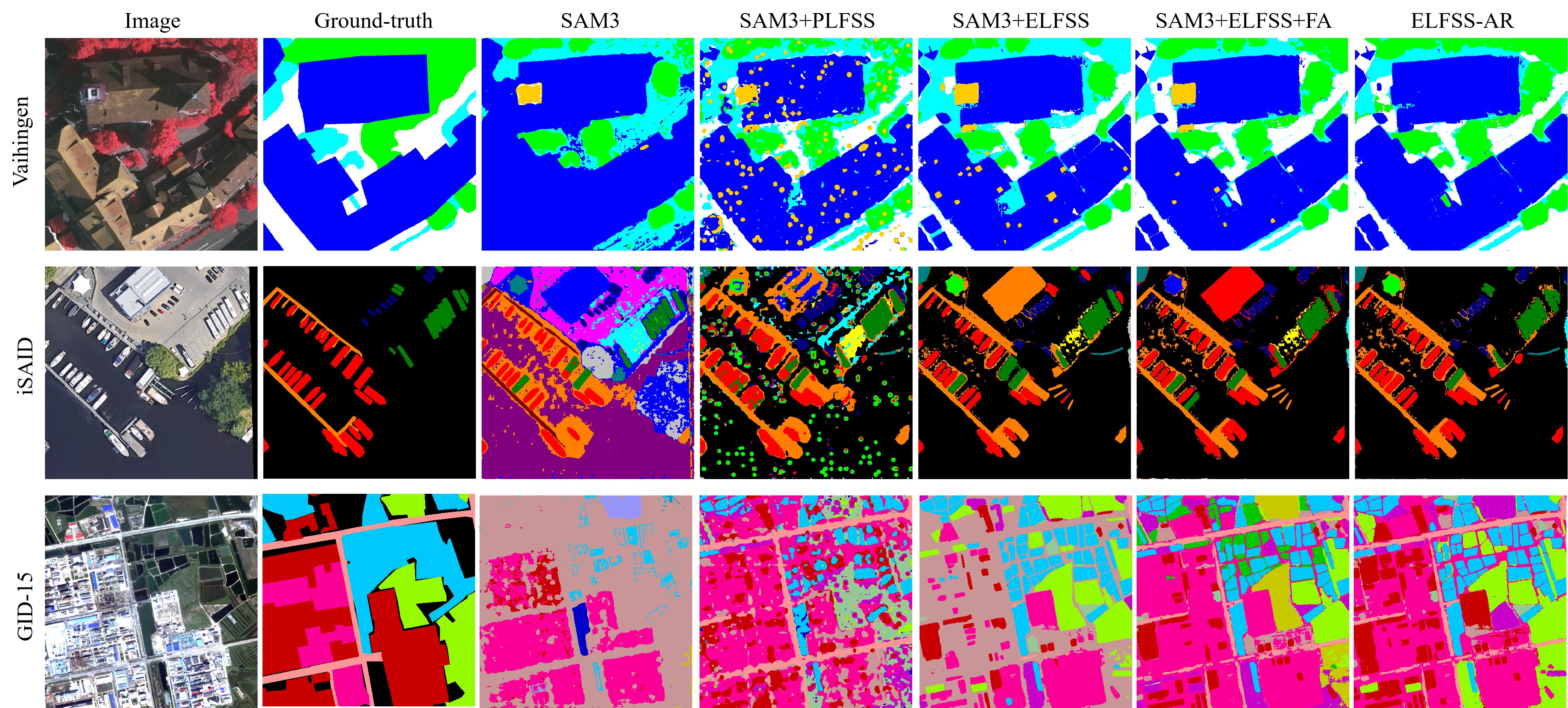}
\caption{Visualization of the ablation study under the 5-shot setting.}
\label{fig:ablation}
\end{figure*}

To provide a more realistic evaluation of the proposed framework, we compare ELFSS-AR with five representative categories of methods, including conventional FSS, CDFSS, foundation-based FSS, training-free OVSS, and support-assisted OVSS. As reported in Table~\ref{tab:pascal_cdfss} and Table~\ref{tab:training_free_ovss}, ELFSS-AR consistently achieves substantially higher segmentation accuracy than all competing approaches. 

Training-free OVSS methods can achieve good performance when a dataset's class names correspond to concepts commonly encountered in general scenes. For example, the Potsdam and Vaihingen datasets include familiar classes such as \emph{building} and \emph{tree}, whose semantics can be effectively captured by pretrained foundation models. However, on datasets such as GID-15 and FBP, which contain domain-specific categories and fine-grained semantic classes, their performance degrades substantially. In particular, SAM3 achieves even lower accuracy than conventional CDFSS methods trained on only a single small-scale dataset.

ConceptBank further leverages target-domain support images to construct a concept bank, alleviating the distribution shift and category-semantic mismatch of frozen SAM3 in the target domain; however, its performance remains relatively limited. In contrast, ELFSS-AR consistently achieves significantly higher accuracy on these challenging datasets. Furthermore, the qualitative results in Fig.~\ref{fig:qualitative_results} under the 5-shot setting show that ELFSS-AR produces complete object regions with spatially coherent internal responses.

\subsection{Ablation Study}
\label{sec:exp_abs}
\begin{figure}[t]
\centering
\includegraphics[width=0.46\textwidth]{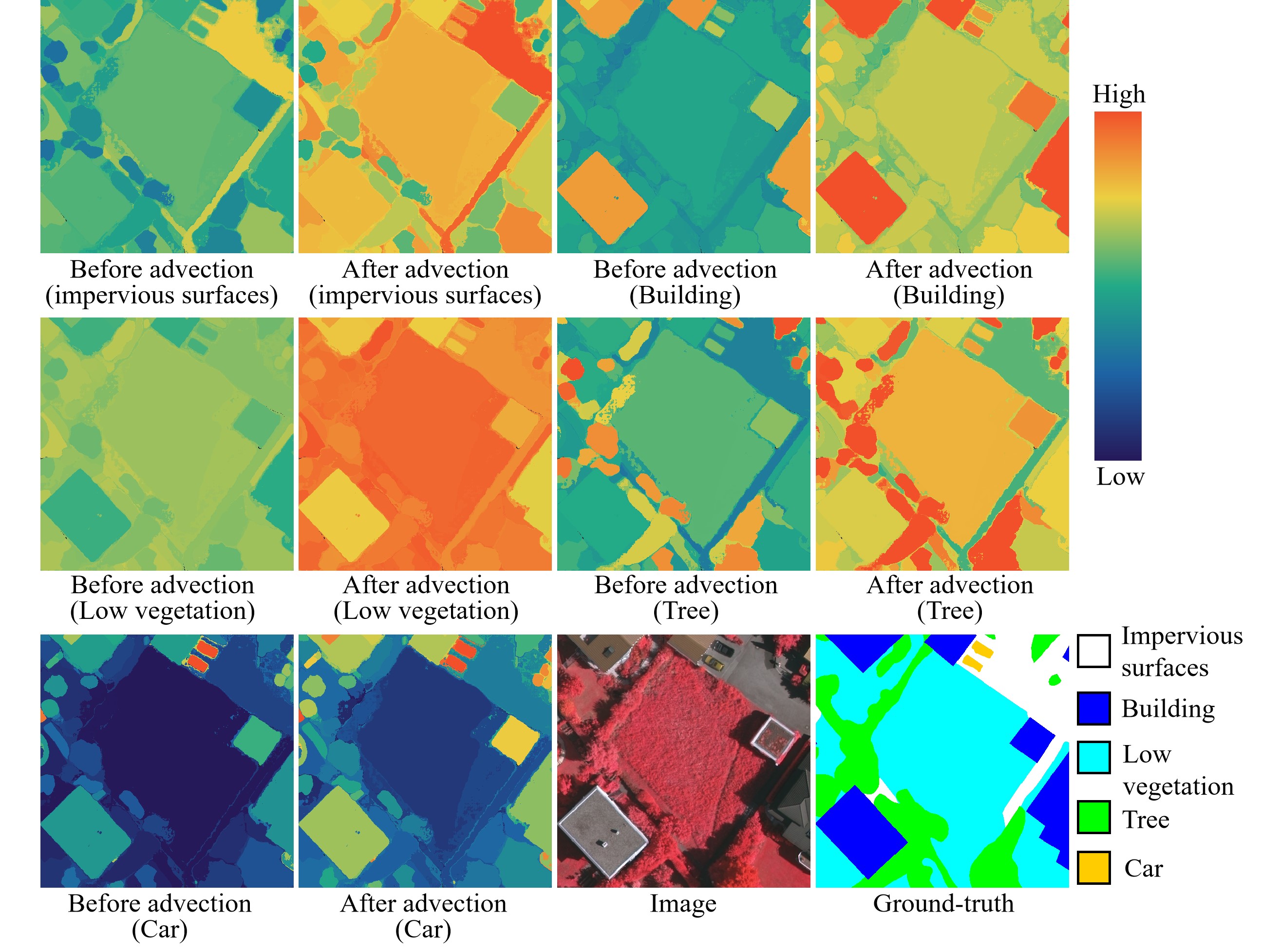}
\caption{Category-wise feature maps before and after advection refinement under the 5-shot setting.}
\label{fig:ablation_fts}
\end{figure}

Table~\ref{fig:ablation} presents the ablation study of all the components of ELFSS-AR. SAM3 is adopted as the baseline. To demonstrate the effectiveness of the proposed entity-level FSS, we further construct a conventional pixel-level FSS on top of SAM3, as reported in the second row of Table~\ref{fig:ablation}. The last row corresponds to the complete ELFSS-AR. It can be observed that each proposed component consistently improves the segmentation performance when incorporated into the baseline. Specifically, as shown in the second and third rows, entity-level FSS outperforms its pixel-level counterpart by 9.57, 4.10 and 2.52 percentage points according to mIoU scores. As shown in the third and last rows, semantic advection refinement brings mIoU improvements of 3.93, 2.21 and 3.09 percentage points. As shown in Fig.~\ref{fig:ablation} and Fig.~\ref{fig:ablation_fts}, the primary purpose of semantic advection is not to further alter the global category decisions of the predicted regions, but to propagate reliable high-confidence semantic information toward local low-confidence regions, thereby correcting intra-class noisy responses and local holes. The detailed theoretical proof is provided in the \textit{Appendix}.

\begin{table}[t]
\centering
\fontsize{6.5pt}{7.5pt}\selectfont
\renewcommand{\arraystretch}{0.9}
\caption{Ablation study of different components under the 5-shot setting. Both mIoU and Boundary IoU (BIoU) are reported. The best results are highlighted in bold, and standard deviations are shown in blue.}
\label{tab:ablation_boundary}
\begin{tabular}{lcccccc}
\toprule
\multirow{2}{*}{Configuration} & \multicolumn{2}{c}{Vaihingen} & \multicolumn{2}{c}{iSAID} & \multicolumn{2}{c}{GID-15}\\
\cmidrule(lr){2-3}\cmidrule(lr){4-5}\cmidrule(lr){6-7}
& mIoU & BIoU & mIoU & BIoU & mIoU & BIoU\\
\midrule
SAM3 & 38.28 & 30.98 & 6.37 & 4.17 & 8.33 & 5.77\\
SAM3+PLFSS
& \makecell{49.87 \\ \textcolor{blue}{\(\pm\) 2.43}}
& \makecell{28.73 \\ \textcolor{blue}{\(\pm\) 1.68}}
& \makecell{16.27 \\ \textcolor{blue}{\(\pm\) 2.49}}
& \makecell{7.24 \\ \textcolor{blue}{\(\pm\) 0.43}}
& \makecell{20.55 \\ \textcolor{blue}{\(\pm\) 2.92}}
& \makecell{6.42 \\ \textcolor{blue}{\(\pm\) 1.36}}\\
SAM3+ELFSS
& \makecell{59.44 \\ \textcolor{blue}{\(\pm\) 2.18}}
& \makecell{40.48 \\ \textcolor{blue}{\(\pm\) 1.65}}
& \makecell{20.37 \\ \textcolor{blue}{\(\pm\) 2.40}}
& \makecell{8.32 \\ \textcolor{blue}{\(\pm\) 0.58}}
& \makecell{23.07 \\ \textcolor{blue}{\(\pm\) 2.57}}
& \makecell{10.82 \\ \textcolor{blue}{\(\pm\) 1.32}}\\
SAM3+ELFSS+FA
& \makecell{61.15 \\ \textcolor{blue}{\(\pm\) 2.10}}
& \makecell{41.96 \\ \textcolor{blue}{\(\pm\) 1.63}}
& \makecell{21.24 \\ \textcolor{blue}{\(\pm\) 2.25}}
& \makecell{8.83 \\ \textcolor{blue}{\(\pm\) 0.52}}
& \makecell{24.19 \\ \textcolor{blue}{\(\pm\) 2.51}}
& \makecell{12.84 \\ \textcolor{blue}{\(\pm\) 1.14}}\\
\textbf{ELFSS-AR}
& \makecell{\textbf{63.37} \\ \textcolor{blue}{\(\pm\) 2.05}}
& \makecell{\textbf{43.10} \\ \textcolor{blue}{\(\pm\) 1.57}}
& \makecell{\textbf{22.58} \\ \textcolor{blue}{\(\pm\) 2.18}}
& \makecell{\textbf{9.92} \\ \textcolor{blue}{\(\pm\) 0.50}}
& \makecell{\textbf{26.16} \\ \textcolor{blue}{\(\pm\) 2.49}}
& \makecell{\textbf{13.97} \\ \textcolor{blue}{\(\pm\) 1.05}}\\
\bottomrule
\end{tabular}
\end{table}

\subsubsection{Entity-Level FSS Enhances Regional Structural Consistency}
By exploiting the generic region modeling capability of SAM3, ELFSS-AR partitions the query image into a set of category-agnostic entity primitives which replace individual pixels as the basic units for inference. As illustrated in Fig.~\ref{fig:ablation}, pixel-level prediction tends to produce local false responses and fragmented segmentation in regions with complex textures. In contrast, introducing entity primitives yields substantially more coherent predictions within the same object, significantly reducing internal holes and fragmented artifacts. Specifically, in the fourth column, the pixel-level FSS misclassifies local texture variations, such as building chimneys and shadows, as vehicles, accompanied by scattered noisy predictions. These issues are effectively alleviated by the proposed entity-level FSS, as demonstrated in the fifth column. 

\subsubsection{Feature Advection Strengthens Category-Aware Representations}
Figure~\ref{fig:ablation_fts} shows the category-wise feature maps before and after query feature advection refinement. Visually, the feature responses across all categories show an overall upward trend after advection refinement. This trend can be attributed to feature transport governed by the two-dimensional advection equation. In this procedure, features belonging to the same object mutually reinforce one another, whereas responses inconsistent with the foreground category are gradually suppressed.

\begin{figure}[t]
\centering
\includegraphics[width=0.48\textwidth]{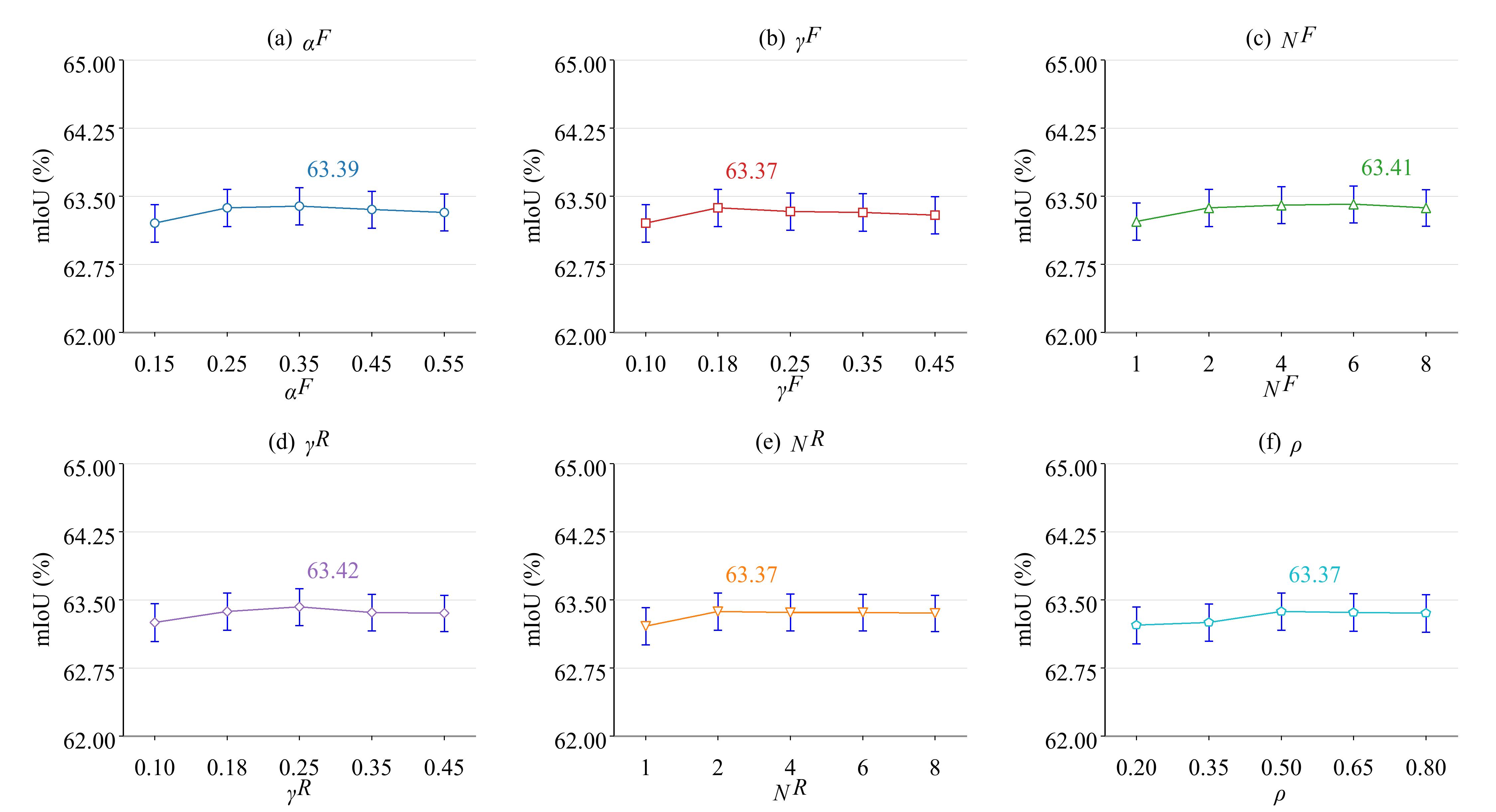}
\caption{Hyperparameter sensitivity analysis on the Vaihingen dataset under the 5-shot setting. Error bars show standard deviations scaled by $0.1$.}
\label{fig:parameter}
\end{figure}

\subsubsection{Response Advection Stabilizes Semantic Inference}
The semantic advection is further applied to the multi-source similarity response space, allowing high-confidence semantic responses to propagate toward neighboring regions along the semantic potential field while suppressing isolated false activations. As illustrated in the last two columns of Fig.~\ref{fig:ablation}, response advection produces substantially more coherent semantic responses within the same entity, correcting local noise and category ambiguities in the predictions.

\subsection{Hyperparameter Sensitivity Analysis}
\label{sec:hyper}
To investigate the effects of key hyperparameters on model performance, we evaluate each hyperparameter while keeping the remaining settings fixed. As shown in Fig.~\ref{fig:parameter}, the model maintains relatively stable performance across the tested ranges. For feature-level textual semantic modulation strength, the highest mIoU of 63.39\% is achieved at $\alpha^F=0.35$. For semantic advection, moderate advection strengths and iteration counts yield better performance, with $\gamma^F=0.18$, $N^F=6$, $\gamma^R=0.25$, and $N^R=2$ achieving favorable results in their respective sweeps. Meanwhile, the model is relatively insensitive to retention weight $\rho$, exhibiting only minor performance variations over a broad range. Overall, ELFSS-AR demonstrates good robustness to its key hyperparameter settings, and we do not require a dedicated tuning procedure to select the default hyperparameter settings. Therefore, the default hyperparameter values reported in the Methods section are not selected to maximize accuracy through extensive hyperparameter sensitivity analyses.

\section{Conclusion}
\label{sec:conclusion}
In this paper, we presented ELFSS-AR, a training-free method built upon SAM3 for CDFSS. ELFSS-AR extends conventional pixel-level prediction to entity-level semantic inference and introduces advection equations to refine both the feature and response spaces. By exploiting only a few labeled support samples from the target domain together with class-specific text prompts, the proposed method effectively segments target-domain images without additional model training. Extensive experiments demonstrate that ELFSS-AR significantly outperforms existing FSS, CDFSS, and training-free OVSS methods, validating the effectiveness of both entity-level FSS and semantic advection refinement. Despite its superior performance, ELFSS-AR still leaves room for further improvement. In particular, the reliability of entity primitive construction could be further enhanced. Future work will therefore investigate more robust entity generation strategies to improve the stability and generalization capability of the proposed method.

\bibliography{ELFSS-AR}

@article{Pascal-2015,
	author = {Everingham, Mark and Eslami, S. M. Ali and Van Gool, Luc and Williams, Christopher K. I. and Winn, John and Zisserman, Andrew},
	date = {2015/01/01},
	doi = {10.1007/s11263-014-0733-5},
	id = {Everingham2015},
	isbn = {1573-1405},
	journal = {International Journal of Computer Vision},
	number = {1},
	pages = {98--136},
	title = {The Pascal Visual Object Classes Challenge: A Retrospective},
	url = {https://doi.org/10.1007/s11263-014-0733-5},
	volume = {111},
	year = {2015}}

@article{ISPRSDATA-2014,
	author = {Franz Rottensteiner and Gunho Sohn and Markus Gerke and Jan Dirk Wegner and Uwe Breitkopf and Jaewook Jung},
	doi = {https://doi.org/10.1016/j.isprsjprs.2013.10.004},
	issn = {0924-2716},
	journal = {ISPRS Journal of Photogrammetry and Remote Sensing},
	pages = {256-271},
	title = {Results of the ISPRS benchmark on urban object detection and 3D building reconstruction},
	url = {https://www.sciencedirect.com/science/article/pii/S0924271613002268},
	volume = {93},
	year = {2014}}

@article{TONG-2023,
	author = {Xin-Yi Tong and Gui-Song Xia and Xiao Xiang Zhu},
	doi = {https://doi.org/10.1016/j.isprsjprs.2022.12.011},
	issn = {0924-2716},
	journal = {ISPRS Journal of Photogrammetry and Remote Sensing},
	pages = {178-196},
	title = {Enabling country-scale land cover mapping with meter-resolution satellite imagery},
	url = {https://www.sciencedirect.com/science/article/pii/S0924271622003264},
	volume = {196},
	year = {2023}}

@article{TONG-2020,
	author = {Xin-Yi Tong and Gui-Song Xia and Qikai Lu and Huanfeng Shen and Shengyang Li and Shucheng You and Liangpei Zhang},
	doi = {https://doi.org/10.1016/j.rse.2019.111322},
	issn = {0034-4257},
	journal = {Remote Sensing of Environment},
	pages = {111322},
	title = {Land-cover classification with high-resolution remote sensing images using transferable deep models},
	url = {https://www.sciencedirect.com/science/article/pii/S0034425719303414},
	volume = {237},
	year = {2020}}

@inproceedings{ProxyCLIP-2025,
	address = {Cham},
	author = {Lan, Mengcheng and Chen, Chaofeng and Ke, Yiping and Wang, Xinjiang and Feng, Litong and Zhang, Wayne},
	booktitle = {Computer Vision -- ECCV 2024},
	editor = {Leonardis, Ale{\v{s}} and Ricci, Elisa and Roth, Stefan and Russakovsky, Olga and Sattler, Torsten and Varol, G{\"u}l},
	isbn = {978-3-031-73113-6},
	pages = {70--88},
	publisher = {Springer Nature Switzerland},
	title = {ProxyCLIP: Proxy Attention Improves CLIP for Open-Vocabulary Segmentation},
	year = {2025}}

@inproceedings{Shao-2025,
	address = {Cham},
	author = {Shao, Tong and Tian, Zhuotao and Zhao, Hang and Su, Jingyong},
	booktitle = {Computer Vision -- ECCV 2024},
	editor = {Leonardis, Ale{\v{s}} and Ricci, Elisa and Roth, Stefan and Russakovsky, Olga and Sattler, Torsten and Varol, G{\"u}l},
	isbn = {978-3-031-73016-0},
	pages = {139--156},
	publisher = {Springer Nature Switzerland},
	title = {Explore the Potential of CLIP for Training-Free Open Vocabulary Semantic Segmentation},
	year = {2025}}

@inproceedings{ClearCLIP-2025,
	address = {Cham},
	author = {Lan, Mengcheng and Chen, Chaofeng and Ke, Yiping and Wang, Xinjiang and Feng, Litong and Zhang, Wayne},
	booktitle = {Computer Vision -- ECCV 2024},
	editor = {Leonardis, Ale{\v{s}} and Ricci, Elisa and Roth, Stefan and Russakovsky, Olga and Sattler, Torsten and Varol, G{\"u}l},
	isbn = {978-3-031-72970-6},
	pages = {143--160},
	publisher = {Springer Nature Switzerland},
	title = {ClearCLIP: Decomposing CLIP Representations for Dense Vision-Language Inference},
	year = {2025}}

@inproceedings{SCLIP-2025,
	address = {Cham},
	author = {Wang, Feng and Mei, Jieru and Yuille, Alan},
	booktitle = {Computer Vision -- ECCV 2024},
	editor = {Leonardis, Ale{\v{s}} and Ricci, Elisa and Roth, Stefan and Russakovsky, Olga and Sattler, Torsten and Varol, G{\"u}l},
	isbn = {978-3-031-72664-4},
	pages = {315--332},
	publisher = {Springer Nature Switzerland},
	title = {SCLIP: Rethinking Self-Attention for Dense Vision-Language Inference},
	year = {2025}}

@inproceedings{Zhou-2022,
	address = {Cham},
	author = {Zhou, Chong and Loy, Chen Change and Dai, Bo},
	booktitle = {Computer Vision -- ECCV 2022},
	editor = {Avidan, Shai and Brostow, Gabriel and Ciss{\'e}, Moustapha and Farinella, Giovanni Maria and Hassner, Tal},
	isbn = {978-3-031-19815-1},
	pages = {696--712},
	publisher = {Springer Nature Switzerland},
	title = {Extract Free Dense Labels from CLIP},
	year = {2022}}

@inproceedings{Li-2025,
	address = {Cham},
	author = {Li, Feng and Zhang, Hao and Sun, Peize and Zou, Xueyan and Liu, Shilong and Li, Chunyuan and Yang, Jianwei and Zhang, Lei and Gao, Jianfeng},
	booktitle = {Computer Vision -- ECCV 2024},
	editor = {Leonardis, Ale{\v{s}} and Ricci, Elisa and Roth, Stefan and Russakovsky, Olga and Sattler, Torsten and Varol, G{\"u}l},
	isbn = {978-3-031-73195-2},
	pages = {467--484},
	publisher = {Springer Nature Switzerland},
	title = {Segment and Recognize Anything at Any Granularity},
	year = {2025}}

@inproceedings{Shaban-2017,
	articleno = {167},
	author = {Shaban, Amirreza and Bansal, Shray and Liu, Zhen and Essa, Irfan and Boots, Byron},
	booktitle = {Proceedings of the British Machine Vision Conference (BMVC)},
	doi = {10.5244/C.31.167},
	editor = {Tae-Kyun Kim, Stefanos Zafeiriou, Gabriel Brostow and Krystian Mikolajczyk},
	isbn = {1-901725-60-X},
	month = {September},
	numpages = {13},
	pages = {167.1-167.13},
	publisher = {BMVA Press},
	title = {One-Shot Learning for Semantic Segmentation},
	url = {https://dx.doi.org/10.5244/C.31.167},
	year = {2017}}

@inproceedings{PANet-2019,
	author = {Wang, Kaixin and Liew, Jun Hao and Zou, Yingtian and Zhou, Daquan and Feng, Jiashi},
	booktitle = {2019 IEEE/CVF International Conference on Computer Vision (ICCV)},
	doi = {10.1109/ICCV.2019.00929},
	pages = {9196-9205},
	title = {PANet: Few-Shot Image Semantic Segmentation With Prototype Alignment},
	year = {2019}}

@article{PFENet-2022,
	author = {Tian, Zhuotao and Zhao, Hengshuang and Shu, Michelle and Yang, Zhicheng and Li, Ruiyu and Jia, Jiaya},
	doi = {10.1109/TPAMI.2020.3013717},
	journal = {IEEE Transactions on Pattern Analysis and Machine Intelligence},
	number = {2},
	pages = {1050-1065},
	title = {Prior Guided Feature Enrichment Network for Few-Shot Segmentation},
	volume = {44},
	year = {2022}}

@article{PFENet-2024,
	author = {Luo, Xiaoliu and Tian, Zhuotao and Zhang, Taiping and Yu, Bei and Tang, Yuan Yan and Jia, Jiaya},
	doi = {10.1109/TPAMI.2023.3329725},
	journal = {IEEE Transactions on Pattern Analysis and Machine Intelligence},
	number = {2},
	pages = {1273-1289},
	title = {PFENet++: Boosting Few-Shot Semantic Segmentation With the Noise-Filtered Context-Aware Prior Mask},
	volume = {46},
	year = {2024}}

@inproceedings{SSP-2022,
	address = {Cham},
	author = {Fan, Qi and Pei, Wenjie and Tai, Yu-Wing and Tang, Chi-Keung},
	booktitle = {Computer {Vision} -- {ECCV} 2022},
	editor = {Avidan, Shai and Brostow, Gabriel and Ciss{\'e}, Moustapha and Farinella, Giovanni Maria and Hassner, Tal},
	isbn = {978-3-031-19800-7},
	pages = {701--719},
	publisher = {Springer Nature Switzerland},
	title = {Self-support {Few}-{Shot} {Semantic} {Segmentation}},
	year = {2022}}

@inproceedings{Peng-2023,
	author = {Peng, Bohao and Tian, Zhuotao and Wu, Xiaoyang and Wang, Chengyao and Liu, Shu and Su, Jingyong and Jia, Jiaya},
	booktitle = {2023 IEEE/CVF Conference on Computer Vision and Pattern Recognition (CVPR)},
	doi = {10.1109/CVPR52729.2023.02264},
	pages = {23641-23651},
	title = {Hierarchical Dense Correlation Distillation for Few-Shot Segmentation},
	year = {2023}}

@inproceedings{DSV-LFS-2025,
	author = {Karimi, Amin and Poullis, Charalambos},
	booktitle = {2025 IEEE/CVF Conference on Computer Vision and Pattern Recognition (CVPR)},
	doi = {10.1109/CVPR52734.2025.00432},
	pages = {4584-4594},
	title = {DSV-LFS: Unifying LLM-Driven Semantic Cues with Visual Features for Robust Few-Shot Segmentation},
	year = {2025}}

@inproceedings{Zou-2025,
	author = {Zou, Tianyu and Xiong, Shengwu and Yao, Ruilin and Rong, Yi},
	booktitle = {2025 IEEE/CVF International Conference on Computer Vision (ICCV)},
	doi = {10.1109/ICCV51701.2025.01912},
	pages = {20561-20571},
	title = {Balancing Conservatism and Aggressiveness: Prototype-Affinity Hybrid Network for Few-Shot Segmentation},
	year = {2025}}

@inproceedings{Lang-2022,
	author = {Lang, Chunbo and Cheng, Gong and Tu, Binfei and Han, Junwei},
	booktitle = {2022 IEEE/CVF Conference on Computer Vision and Pattern Recognition (CVPR)},
	doi = {10.1109/CVPR52688.2022.00789},
	pages = {8047-8057},
	title = {Learning What Not to Segment: A New Perspective on Few-Shot Segmentation},
	year = {2022}}

@article{Lu-2023,
	author = {Lu, Zhihe and He, Sen and Li, Da and Song, Yi-Zhe and Xiang, Tao},
	doi = {10.1109/TIP.2023.3282070},
	journal = {IEEE Transactions on Image Processing},
	pages = {3311-3323},
	title = {Prediction Calibration for Generalized Few-Shot Semantic Segmentation},
	volume = {32},
	year = {2023}}

@inproceedings{SAM-2023,
	author = {Kirillov, Alexander and Mintun, Eric and Ravi, Nikhila and Mao, Hanzi and Rolland, Chloe and Gustafson, Laura and Xiao, Tete and Whitehead, Spencer and Berg, Alexander C. and Lo, Wan-Yen and Doll{\'a}r, Piotr and Girshick, Ross},
	booktitle = {2023 IEEE/CVF International Conference on Computer Vision (ICCV)},
	doi = {10.1109/ICCV51070.2023.00371},
	pages = {3992-4003},
	title = {Segment Anything},
	year = {2023}}

@inproceedings{Xu-2025,
	author = {Qianxiong Xu and Lanyun Zhu and Xuanyi Liu and Guosheng Lin and Cheng Long and Ziyue Li and Rui Zhao},
	booktitle = {Forty-second International Conference on Machine Learning},
	title = {Unlocking the Power of {SAM} 2 for Few-Shot Segmentation},
	url = {https://openreview.net/forum?id=TQtUTC3eKv},
	year = {2025}}

@inproceedings{CDFSS-Lei-2022,
	address = {Cham},
	author = {Lei, Shuo and Zhang, Xuchao and He, Jianfeng and Chen, Fanglan and Du, Bowen and Lu, Chang-Tien},
	booktitle = {Computer Vision -- ECCV 2022},
	editor = {Avidan, Shai and Brostow, Gabriel and Ciss{\'e}, Moustapha and Farinella, Giovanni Maria and Hassner, Tal},
	isbn = {978-3-031-20056-4},
	pages = {73--90},
	publisher = {Springer Nature Switzerland},
	title = {Cross-Domain Few-Shot Semantic Segmentation},
	year = {2022}}

@inproceedings{CDFSS-Herzog2024,
	author = {Herzog, Jonas},
	booktitle = {2024 IEEE/CVF Conference on Computer Vision and Pattern Recognition (CVPR)},
	doi = {10.1109/CVPR52733.2024.02228},
	pages = {23605-23615},
	title = {Adapt Before Comparison: A New Perspective on Cross-Domain Few-Shot Segmentation},
	year = {2024}}

@inproceedings{CDFSS-Su2024,
	author = {Su, Jiapeng and Fan, Qi and Pei, Wenjie and Lu, Guangming and Chen, Fanglin},
	booktitle = {2024 IEEE/CVF Conference on Computer Vision and Pattern Recognition (CVPR)},
	doi = {10.1109/CVPR52733.2024.02269},
	pages = {24036-24045},
	title = {Domain-Rectifying Adapter for Cross-Domain Few-Shot Segmentation},
	year = {2024}}

@inproceedings{CDFSS-Nie2024,
	author = {Nie, Jiahao and Xing, Yun and Zhang, Gongjie and Yan, Pei and Xiao, Aoran and Tan, Yap-Peng and Kot, Alex C. and Lu, Shijian},
	booktitle = {2024 IEEE/CVF Conference on Computer Vision and Pattern Recognition (CVPR)},
	doi = {10.1109/CVPR52733.2024.00325},
	pages = {3380-3390},
	title = {Cross-Domain Few-Shot Segmentation via Iterative Support-Query Correspondence Mining},
	year = {2024}}

@inproceedings{CDFSS-He-2024,
	author = {He, Weizhao and Zhang, Yang and Zhuo, Wei and Shen, Linlin and Yang, Jiaqi and Deng, Songhe and Sun, Liang},
	booktitle = {2024 IEEE/CVF Conference on Computer Vision and Pattern Recognition (CVPR)},
	doi = {10.1109/CVPR52733.2024.02243},
	pages = {23762-23772},
	title = {APSeg: Auto-Prompt Network for Cross-Domain Few-Shot Semantic Segmentation},
	year = {2024}}

@inproceedings{Shi-2026,
	author = {Shi, Guangchen and Wu, Yirui and Zhu, Wei and Wang, Tao and Zhang, Hao and Li, Bo and Lu, Tong},
	booktitle = {Proceedings of the IEEE/CVF Conference on Computer Vision and Pattern Recognition (CVPR)},
	date = {2026-05-05},
	month = {June},
	pages = {12354-12363},
	title = {Bayesian Decomposition and Semantic Completion for Few-shot Semantic Segmentation},
	url = {https://openaccess.thecvf.com/content/CVPR2026/html/Shi_Bayesian_Decomposition_and_Semantic_Completion_for_Few-shot_Semantic_Segmentation_CVPR_2026_paper.html},
	year = {2026}}

@inproceedings{CDFSS-Li-2025,
	author = {Li, Zhaoyang and Wang, Yuan and Li, Wangkai and Zhang, Tianzhu and Liu, Xiang},
	booktitle = {2025 IEEE/CVF Conference on Computer Vision and Pattern Recognition (CVPR)},
	doi = {10.1109/CVPR52734.2025.00920},
	pages = {9849-9859},
	title = {Dual-Agent Optimization framework for Cross-Domain Few-Shot Segmentation},
	year = {2025}}

@inproceedings{CDFSS-Liu-2025,
	author = {Liu, Yuhan and Zou, Yixiong and Li, Yuhua and Li, Ruixuan},
	booktitle = {2025 IEEE/CVF Conference on Computer Vision and Pattern Recognition (CVPR)},
	doi = {10.1109/CVPR52734.2025.00435},
	pages = {4618-4627},
	title = {The Devil is in Low-Level Features for Cross-Domain Few-Shot Segmentation},
	year = {2025}}

@inproceedings{CDFSS-Tong-2025,
	author = {Jintao Tong and Ran Ma and Yixiong Zou and Guangyao Chen and Yuhua Li and Ruixuan Li},
	booktitle = {Forty-second International Conference on Machine Learning},
	title = {Adapter Naturally Serves as Decoupler for Cross-Domain Few-Shot Semantic Segmentation},
	url = {https://openreview.net/forum?id=Pokj70ZAxJ},
	year = {2025}}

@inproceedings{SAM2-2025,
	author = {Nikhila Ravi and Valentin Gabeur and Yuan-Ting Hu and Ronghang Hu and Chaitanya Ryali and Tengyu Ma and Haitham Khedr and Roman R{\"a}dle and Chloe Rolland and Laura Gustafson and Eric Mintun and Junting Pan and Kalyan Vasudev Alwala and Nicolas Carion and Chao-Yuan Wu and Ross Girshick and Piotr Dollar and Christoph Feichtenhofer},
	booktitle = {The Thirteenth International Conference on Learning Representations},
	title = {{SAM} 2: Segment Anything in Images and Videos},
	url = {https://openreview.net/forum?id=Ha6RTeWMd0},
	year = {2025}}

@inproceedings{Zou-2023b,
	author = {Xueyan Zou and Jianwei Yang and Hao Zhang and Feng Li and Linjie Li and Jianfeng Wang and Lijuan Wang and Jianfeng Gao and Yong Jae Lee},
	booktitle = {Thirty-seventh Conference on Neural Information Processing Systems},
	title = {Segment Everything Everywhere All at Once},
	url = {https://openreview.net/forum?id=UHBrWeFWlL},
	year = {2023}}

@inproceedings{Zhang-2023,
	author = {Zhang, Hao and Li, Feng and Zou, Xueyan and Liu, Shilong and Li, Chunyuan and Yang, Jianwei and Zhang, Lei},
	booktitle = {2023 IEEE/CVF International Conference on Computer Vision (ICCV)},
	doi = {10.1109/ICCV51070.2023.00100},
	pages = {1020-1031},
	title = {A Simple Framework for Open-Vocabulary Segmentation and Detection},
	year = {2023}}

@inproceedings{Liu-2025,
	author = {Liu, Yang and Yin, Yufei and Jing, Chenchen and Zhu, Muzhi and Chen, Hao and Xi, Yuling and Feng, Bo and Wang, Hao and Li, Shiyu and Shen, Chunhua},
	booktitle = {2025 IEEE/CVF International Conference on Computer Vision (ICCV)},
	doi = {10.1109/ICCV51701.2025.02002},
	pages = {21557-21567},
	title = {Unified Open-World Segmentation with Multi-Modal Prompts},
	year = {2025}}

@inproceedings{LISA-2024,
	author = {Lai, Xin and Tian, Zhuotao and Chen, Yukang and Li, Yanwei and Yuan, Yuhui and Liu, Shu and Jia, Jiaya},
	booktitle = {2024 IEEE/CVF Conference on Computer Vision and Pattern Recognition (CVPR)},
	doi = {10.1109/CVPR52733.2024.00915},
	pages = {9579-9589},
	title = {LISA: Reasoning Segmentation via Large Language Model},
	year = {2024}}

@inproceedings{GLaMM-2024,
	author = {Rasheed, Hanoona and Maaz, Muhammad and Shaji, Sahal and Shaker, Abdelrahman and Khan, Salman and Cholakkal, Hisham and Anwer, Rao M. and Xing, Eric and Yang, Ming-Hsuan and Khan, Fahad S.},
	booktitle = {2024 IEEE/CVF Conference on Computer Vision and Pattern Recognition (CVPR)},
	doi = {10.1109/CVPR52733.2024.01236},
	pages = {13009-13018},
	title = {GLaMM: Pixel Grounding Large Multimodal Model},
	year = {2024}}

@inproceedings{PixelLM-2024,
	author = {Ren, Zhongwei and Huang, Zhicheng and Wei, Yunchao and Zhao, Yao and Fu, Dongmei and Feng, Jiashi and Jin, Xiaojie},
	booktitle = {2024 IEEE/CVF Conference on Computer Vision and Pattern Recognition (CVPR)},
	doi = {10.1109/CVPR52733.2024.02491},
	pages = {26364-26373},
	title = {PixelLM: Pixel Reasoning with Large Multimodal Model},
	year = {2024}}

@inproceedings{Hajimiri-2025,
	author = {Hajimiri, Sina and Ayed, Ismail Ben and Dolz, Jose},
	booktitle = {2025 IEEE/CVF Winter Conference on Applications of Computer Vision (WACV)},
	doi = {10.1109/WACV61041.2025.00495},
	pages = {5061-5071},
	title = {Pay Attention to Your Neighbours: Training-Free Open-Vocabulary Semantic Segmentation},
	year = {2025}}

@inproceedings{ResCLIP-2025,
	author = {Yang, Yuhang and Deng, Jinhong and Li, Wen and Duan, Lixin},
	booktitle = {2025 IEEE/CVF Conference on Computer Vision and Pattern Recognition (CVPR)},
	doi = {10.1109/CVPR52734.2025.02789},
	pages = {29968-29978},
	title = {ResCLIP: Residual Attention for Training-free Dense Vision-language Inference},
	year = {2025}}

@inproceedings{Shi-2025,
	author = {Shi, Yuheng and Dong, Minjing and Xu, Chang},
	booktitle = {2025 IEEE/CVF International Conference on Computer Vision (ICCV)},
	doi = {10.1109/ICCV51701.2025.02180},
	pages = {23487-23497},
	title = {Harnessing Vision Foundation Models for High-Performance, Training-Free Open Vocabulary Segmentation},
	year = {2025}}

@inproceedings{SegEarth-OV-2025,
	author = {Li, Kaiyu and Liu, Ruixun and Cao, Xiangyong and Bai, Xueru and Zhou, Feng and Meng, Deyu and Wang, Zhi},
	booktitle = {2025 IEEE/CVF Conference on Computer Vision and Pattern Recognition (CVPR)},
	doi = {10.1109/CVPR52734.2025.00986},
	pages = {10545-10556},
	title = {SegEarth-OV: Towards Training-Free Open-Vocabulary Segmentation for Remote Sensing Images},
	year = {2025}}

@article{Snell-2017,
	author = {Snell, Jake and Swersky, Kevin and Zemel, Richard},
	journal = {Advances in neural information processing systems},
	title = {Prototypical networks for few-shot learning},
	volume = {30},
	year = {2017}}

@inproceedings{ReAttnCLIP-2026,
	author = {Niu, Xin and Zhao, Manqi and Jiang, Dongsheng and Wu, Yingying and Su, Bing},
	booktitle = {Proceedings of the IEEE/CVF Conference on Computer Vision and Pattern Recognition (CVPR)},
	month = {June},
	pages = {24980-24989},
	title = {ReAttnCLIP: Training-Free Open-Vocabulary Remote Sensing Image Segmentation via Re-defined Attention in CLIP},
	year = {2026}}

@inproceedings{iSAID-2019,
	author = {Waqas Zamir, Syed and Arora, Aditya and Gupta, Akshita and Khan, Salman and Sun, Guolei and Shahbaz Khan, Fahad and Zhu, Fan and Shao, Ling and Xia, Gui-Song and Bai, Xiang},
	booktitle = {Proceedings of the IEEE/CVF conference on computer vision and pattern recognition workshops},
	pages = {28--37},
	title = {isaid: A large-scale dataset for instance segmentation in aerial images},
	year = {2019}}

@inproceedings{SAM3-2026,
	author = {Nicolas Carion and Laura Gustafson and Yuan-Ting Hu and Shoubhik Debnath and Ronghang Hu and Didac Suris Coll-Vinent and Chaitanya Ryali and Kalyan Vasudev Alwala and Haitham Khedr and Andrew Huang and Jie Lei and Tengyu Ma and Baishan Guo and Arpit Kalla and Markus Marks and Joseph Greer and Meng Wang and Peize Sun and Roman R{\"a}dle and Triantafyllos Afouras and Effrosyni Mavroudi and Katherine Xu and Tsung-Han Wu and Yu Zhou and Liliane Momeni and RISHI HAZRA and Shuangrui Ding and Sagar Vaze and Francois Porcher and Feng Li and Siyuan Li and Aishwarya Kamath and Ho Kei Cheng and Piotr Dollar and Nikhila Ravi and Kate Saenko and Pengchuan Zhang and Christoph Feichtenhofer},
	booktitle = {The Fourteenth International Conference on Learning Representations},
	title = {{SAM} 3: Segment Anything with Concepts},
	url = {https://openreview.net/forum?id=r35clVtGzw},
	year = {2026}}

@article{ICPD-2025,
	author = {Jiang, Zhiyu and Yuan, Ye and Ma, Dandan and Wang, Qi and Yuan, Yuan},
	doi = {10.1109/TGRS.2025.3617662},
	journal = {IEEE Transactions on Geoscience and Remote Sensing},
	pages = {1-13},
	title = {Implicit CLIP Prior Decoupling for Few-Shot Remote Sensing Image Segmentation},
	volume = {63},
	year = {2025}}

@inproceedings{APM-2024,
 author = {Tong, Jintao and Zou, Yixiong and Li, Yuhua and Li, Ruixuan},
 booktitle = {Advances in Neural Information Processing Systems},
 doi = {10.52202/079017-3066},
 editor = {A. Globerson and L. Mackey and D. Belgrave and A. Fan and U. Paquet and J. Tomczak and C. Zhang},
 pages = {96728--96749},
 publisher = {Curran Associates, Inc.},
 title = {Lightweight Frequency Masker for Cross-Domain  Few-Shot Semantic Segmentation},
 url = {https://proceedings.neurips.cc/paper_files/paper/2024/file/af536fa4312b5b2e5c541156cd31e93b-Paper-Conference.pdf},
 volume = {37},
 year = {2024},
}

@Article{FLAIR-2022,
  author     = {Anatol Garioud and St{\'{e}}phane Peillet and Eva Bookjans and S{\'{e}}bastien Giordano and Boris Wattrelos},
  journal    = {CoRR},
  title      = {{FLAIR} {\#}1: semantic segmentation and domain adaptation dataset},
  year       = {2022},
  volume     = {abs/2211.12979},
  doi        = {10.48550/ARXIV.2211.12979},
  eprint     = {2211.12979},
  eprinttype = {arXiv},
  url        = {https://doi.org/10.48550/arXiv.2211.12979},
}

@Data{MiniFrance-2020,
  author    = {Javiera Castillo Navarro and Bertrand Le Saux and Alexandre Boulch and Nicolas Audebert and Sébastien Lefèvre},
  doi       = {10.21227/b9pt-8x03},
  publisher = {IEEE Dataport},
  title     = {MiniFrance},
  url       = {https://dx.doi.org/10.21227/b9pt-8x03},
  year      = {2020},
}

@misc{conceptbank_2026,
      title={Taming SAM3 in the Wild: A Concept Bank for Open-Vocabulary Segmentation}, 
      author={Gensheng Pei and Xiruo Jiang and Yazhou Yao and Xiangbo Shu and Fumin Shen and Byeungwoo Jeon},
      year={2026},
      eprint={2602.06333},
      archivePrefix={arXiv},
      primaryClass={cs.CV},
      url={https://arxiv.org/abs/2602.06333}, 
}

@article{Remoteclip_2024,
  title={Remoteclip: A vision language foundation model for remote sensing},
  author={Liu, Fan and Chen, Delong and Guan, Zhangqingyun and Zhou, Xiaocong and Zhu, Jiale and Ye, Qiaolin and Fu, Liyong and Zhou, Jun},
  journal={IEEE Transactions on Geoscience and Remote Sensing},
  volume={62},
  pages={1--16},
  year={2024},
}

@article{DINOv2_2024,
  title     = {{DINOv2: Learning Robust Visual Features Without Supervision}},
  author    = {Oquab, Maxime and Darcet, Timothée and Moutakanni, Théo and Vo, Huy V. and Szafraniec, Marc and Khalidov, Vasil and Fernandez, Pierre and Haziza, Daniel and Massa, Francisco and El-Nouby, Alaaeldin and Assran, Mido and Ballas, Nicolas and Galuba, Wojciech and Howes, Russell and Huang, Po-Yao and Li, Shang-Wen and Misra, Ishan and Rabbat, Michael and Sharma, Vasu and Synnaeve, Gabriel and Xu, Hu and Jegou, Herve and Mairal, Julien and Labatut, Patrick and Joulin, Armand and Bojanowski, Piotr},
  journal   = {Transactions on Machine Learning Research},
  year      = {2024},
  url       = {https://mlanthology.org/tmlr/2024/oquab2024tmlr-dinov2/},
}

@InProceedings{CLIP_2021,
  title = 	 {Learning Transferable Visual Models From Natural Language Supervision},
  author =       {Radford, Alec and Kim, Jong Wook and Hallacy, Chris and Ramesh, Aditya and Goh, Gabriel and Agarwal, Sandhini and Sastry, Girish and Askell, Amanda and Mishkin, Pamela and Clark, Jack and Krueger, Gretchen and Sutskever, Ilya},
  booktitle = 	 {Proceedings of the 38th International Conference on Machine Learning},
  pages = 	 {8748--8763},
  year = 	 {2021},
  editor = 	 {Meila, Marina and Zhang, Tong},
  volume = 	 {139},
  series = 	 {Proceedings of Machine Learning Research},
  month = 	 {18--24 Jul},
  publisher =    {PMLR},
  url = 	 {https://proceedings.mlr.press/v139/radford21a.html},
}

@InProceedings{BIoU_2021,
    author    = {Cheng, Bowen and Girshick, Ross and Dollar, Piotr and Berg, Alexander C. and Kirillov, Alexander},
    title     = {Boundary IoU: Improving Object-Centric Image Segmentation Evaluation},
    booktitle = {Proceedings of the IEEE/CVF Conference on Computer Vision and Pattern Recognition (CVPR)},
    month     = {June},
    year      = {2021},
    pages     = {15334-15342}
}

\appendix

\section{Details of Category-Agnostic Entity Primitive Generation}
\subsection{Generation}
Remote-sensing objects generally exhibit strong spatial continuity: pixels within the same geographic region tend to share consistent semantic labels, whereas category transitions mainly occur along object boundaries. Accordingly, the semantic label map can be approximately modeled as a piecewise-constant function over spatially continuous regions. Conventional pixel-wise independent prediction is susceptible to local textures, spectral fluctuations, and domain shifts, often resulting in spatially scattered responses within homogeneous regions. Let the prediction response of pixel $x$ for class $c$ be
\begin{equation}
S_c(x)=\mu_c(x)+\epsilon_c(x),
\end{equation}
where $\mu_c(x)$ denotes the underlying semantic response and $\epsilon_c(x)$ represents a local perturbation. For an object region $A$, its region-level response is defined as
\begin{equation}
\bar{S}_c(A)=\frac{1}{|A|}\sum_{x\in A}S_c(x).
\end{equation}
Assuming that the perturbations within the region are approximately independent and have variance $\sigma^2$, we have
\begin{equation}
\operatorname{Var}\!\left[\bar{S}_c(A)\right]=\frac{\sigma^2}{|A|},
\end{equation}
which indicates that region-level aggregation reduces the variance of local perturbations and improves intra-region semantic consistency. Motivated by this observation, we exploit the generic geometric prior of SAM3 to partition the query image $I_q$ into a set of category-agnostic entities that are spatially exclusive and collectively cover the entire image, thereby extending few-shot segmentation from pixel-level inference to entity-level inference.

Given a query image $I_q$, we first employ the automatic mask generation mechanism of SAM3 to perform full-image inference using densely sampled point prompts. This process produces $N$ initial candidate masks together with their corresponding quality scores:
\begin{equation}
\mathcal{R}_q=
\left\{
\left(r_n,s_n^{\mathrm{sam}}\right)
\right\}_{n=1}^{N},
\qquad
r_n\in\{0,1\}^{H\times W},
\end{equation}
where $r_n$ denotes the $n$-th binary candidate mask and $s_n^{\mathrm{sam}}\in\mathbb{R}$ is the corresponding quality score predicted by SAM3. Let $\Omega_q$ denote the pixel domain of the query image. The area of a candidate mask is defined as
\begin{equation}
\operatorname{Area}(r_n)
=
\sum_{x\in\Omega_q}r_n(x).
\end{equation}

Small candidate masks produced during automatic mask generation are often induced by local textures, high-frequency noise, or boundary perturbations. Because such regions contain only a limited number of pixels, their aggregated responses tend to exhibit relatively high variance and low semantic stability. To suppress these trivial candidate regions, we introduce a minimum area threshold $\theta_{\mathrm{area}}=10$ and obtain the filtered candidate set
\begin{equation}
\mathcal{R}_q^{f}
=
\left\{
\left(r_n,s_n^{\mathrm{sam}}\right)\in\mathcal{R}_q
\;\middle|\;
\operatorname{Area}(r_n)\geq\theta_{\mathrm{area}}
\right\}.
\end{equation}
Suppose that $N'$ candidate masks remain after filtering. We sort them in descending order according to their SAM3 quality scores and define a permutation $\pi$ satisfying
\begin{equation}
s_{\pi(1)}^{\mathrm{sam}}
\geq
s_{\pi(2)}^{\mathrm{sam}}
\geq
\cdots
\geq
s_{\pi(N')}^{\mathrm{sam}}.
\end{equation}

Since the candidate masks automatically generated by SAM3 often exhibit substantial containment and overlap, directly classifying these overlapping masks would allow the same pixel to participate in multiple entity-level decisions, leading to ambiguous semantic assignments. We therefore adopt a confidence-prioritized greedy exclusive assignment strategy to transform the overlapping candidate masks into spatially disjoint entity primitives. Let $A^{(j)}\subseteq\Omega_q$ denote the set of pixels that have already been assigned after processing the first $j$ candidate masks, and let $\mathcal{G}^{(j)}$ denote the resulting set of entity primitives. The process is initialized as
\begin{equation}
A^{(0)}=\varnothing,
\qquad
\mathcal{G}^{(0)}=\varnothing.
\end{equation}
For the $j$-th candidate mask $r_{\pi(j)}$ in the sorted sequence, its effective region after removing the already assigned pixels is defined as
\begin{equation}
\hat{r}_{\pi(j)}
=
r_{\pi(j)}\setminus A^{(j-1)}.
\end{equation}
The occupied pixel set and the entity primitive set are then updated according to the area of the effective region:
\begin{equation}
A^{(j)}
=
\begin{cases}
A^{(j-1)}\cup\hat{r}_{\pi(j)},
&
\operatorname{Area}\!\left(\hat{r}_{\pi(j)}\right)
\geq
\theta_{\mathrm{area}},
\\
A^{(j-1)},
&
\operatorname{Area}\!\left(\hat{r}_{\pi(j)}\right)
<
\theta_{\mathrm{area}},
\end{cases}
\end{equation}
and
\begin{equation}
\mathcal{G}^{(j)}
=
\begin{cases}
\mathcal{G}^{(j-1)}
\cup
\left\{\hat{r}_{\pi(j)}\right\},
&
\operatorname{Area}\!\left(\hat{r}_{\pi(j)}\right)
\geq
\theta_{\mathrm{area}},
\\
\mathcal{G}^{(j-1)},
&
\operatorname{Area}\!\left(\hat{r}_{\pi(j)}\right)
<
\theta_{\mathrm{area}}.
\end{cases}
\end{equation}

To control the computational cost, we set the maximum number of entity primitives to $N_{\max}=500$. The above procedure terminates when all candidate masks have been processed or when the number of valid entity primitives reaches $N_{\max}-1$. Let $J$ denote the termination step. The pixels that remain unassigned are grouped into a residual entity primitive:
\begin{equation}
g_{\mathrm{res}}
=
\Omega_q\setminus A^{(J)}.
\end{equation}
The residual primitive participates in the subsequent entity-level semantic inference together with the other entity primitives, and its overall class is determined by aggregating the multi-source semantic responses within the region. The final entity primitive set is therefore defined as
\begin{equation}
\mathcal{G}_q
=
\begin{cases}
\mathcal{G}^{(J)}\cup\{g_{\mathrm{res}}\},
&
\operatorname{Area}(g_{\mathrm{res}})>0,
\\
\mathcal{G}^{(J)},
&
\operatorname{Area}(g_{\mathrm{res}})=0.
\end{cases}
\end{equation}

Consequently, the final set of entity primitives can be written as
\begin{equation}
\mathcal{G}_q=\{g_i\}_{i=1}^{L},
\qquad
L\leq N_{\max}.
\end{equation}
These entity primitives are pairwise disjoint and collectively cover the entire query image:
\begin{equation}
\begin{cases}
g_i\cap g_j=\varnothing,
&
i\neq j,
\\
\displaystyle\bigcup_{i=1}^{L}g_i=\Omega_q.
\end{cases}
\end{equation}
The entity primitives generated from the sorted candidate masks inherit the corresponding SAM3 quality scores $s_n^{\mathrm{sam}}$, $n=1,\dots, N$, whereas the quality score of the residual entity primitive is set to zero. These category-agnostic primitives provide only spatial and geometric support; their semantic labels are determined during the subsequent entity-level semantic inference stage. 

\subsection{Hyperparameter Analysis}
\label{sec:entity_hyper}

\begin{figure}[t]
\centering
\includegraphics[width=0.48\textwidth]{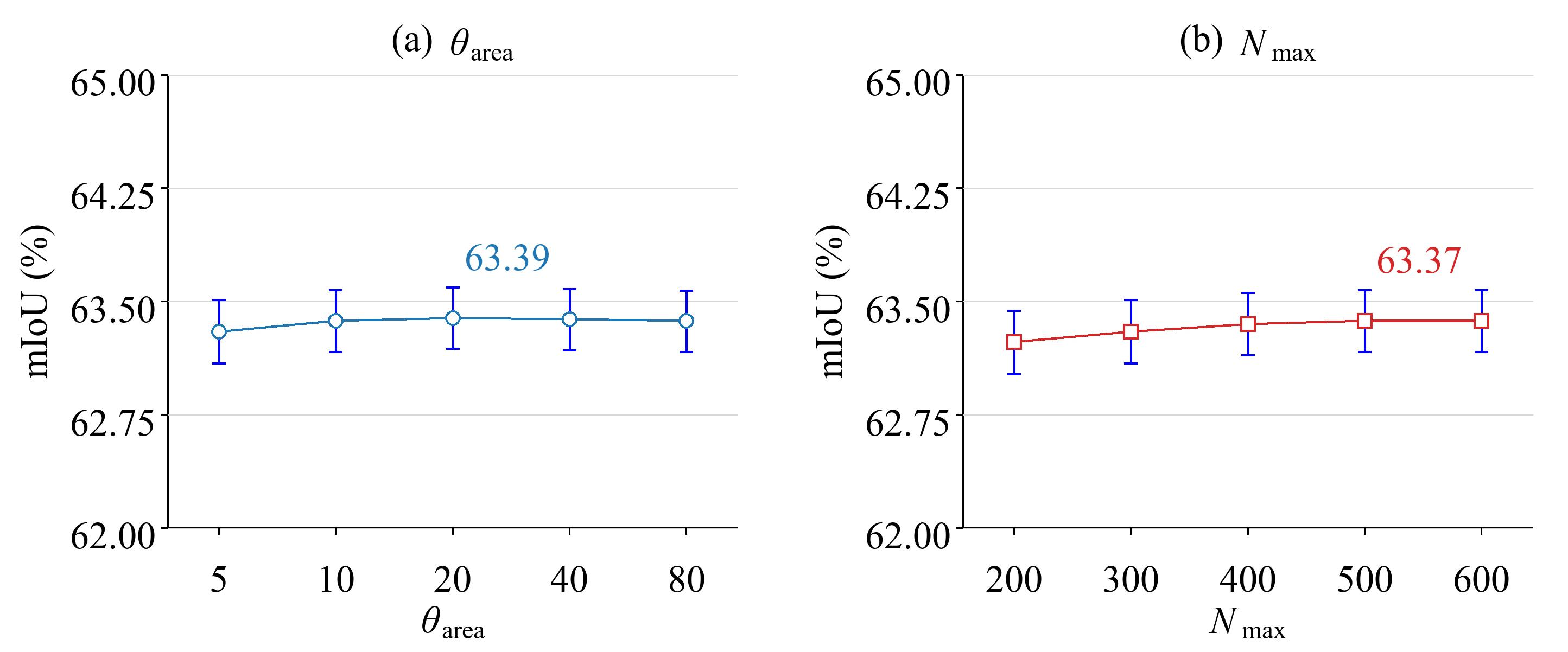}
\caption{Hyperparameter analysis of entity primitive generation on the Vaihingen dataset under the 5-shot setting. Error bars show standard deviations scaled by $0.1$.}
\label{fig:parameter_entity}
\end{figure}

To investigate the effects of hyperparameters in entity primitive generation on model performance, we evaluate each hyperparameter while keeping the remaining settings fixed. As shown in Fig.~\ref{fig:parameter_entity}, the model maintains relatively stable performance across the tested ranges. For the minimum area threshold, the highest mIoU of 63.39\% is achieved at $\theta_{\mathrm{area}}=20$, with minimal performance variations across the tested values. For the maximum number of entity primitives, performance gradually improves as $N_{\max}$ increases, reaching an mIoU of 63.37\% at $N_{\max}=500$. Overall, ELFSS-AR demonstrates good robustness to its entity primitive generation hyperparameters.

{\bf Notably}, the peak accuracy shown in Fig.~\ref{fig:parameter_entity} does not necessarily match the accuracy reported in the comparative experiments. As explained in the main text, to limit the influence of extensive sensitivity analyses on the reported performance, we did not use such analyses to select hyperparameter values. Instead, we adopted default values based on prior experience. All accuracies reported in the comparative experiments were obtained using these default settings. Consequently, when the peak accuracy in Fig.~\ref{fig:parameter_entity} occurs at the default value of the hyperparameter being examined, it matches the accuracy reported in the comparative experiments. Otherwise, it exceeds the reported accuracy. The same applies to the line plots for the sensitivity analyses of the other hyperparameters. 

\section{Experimental Settings}
\subsection{Datasets}
To comprehensively evaluate the performance of the proposed method across different sensors, spatial resolutions, and complex land-cover/land-use scenarios, we conduct experiments on seven representative and challenging remote-sensing semantic segmentation datasets. For all datasets, the official class names are directly used as text prompts.

\textbf{1) GID Dataset.}
The GID dataset~\cite{TONG-2020} is a large-scale high-resolution remote-sensing dataset constructed from Gaofen-2 (GF-2) satellite imagery. Each image has a spatial size of $6800\times7200$ pixels and a spatial resolution of 4~m. GID consists of two subsets: the large-scale classification set, GID-5, and the fine-grained land-cover set, GID-15. GID-5 contains five land-cover categories, including \textit{built-up}, \textit{farmland}, \textit{forest}, \textit{meadow}, and \textit{water}. GID-15 further divides the land-cover types into 15 fine-grained categories, such as \textit{paddy field}, \textit{irrigated land}, \textit{dry cropland}, \textit{garden land}, and \textit{arbor forest}. At a spatial resolution of 4~m, land-cover regions often exhibit substantial intra-class spectral heterogeneity. Consequently, training-free segmentation based on direct feature matching is prone to producing random noise and isolated fragmented predictions within homogeneous regions. This problem is particularly severe on GID-15, where the visual similarity among fine-grained categories is high and the number of classes is relatively large. Under such conditions, multimodal vision foundation models relying on generic text prompts, such as SAM3, generally exhibit limited recognition accuracy. Evaluations on this dataset therefore not only verify the adaptive refinement capability of semantic advection in suppressing feature noise, smoothing land-cover regions, and correcting isolated false predictions, but also demonstrate the domain adaptability obtained by combining the geometric entity extraction capability of SAM3 with few-shot semantic inference in highly confusing fine-grained remote-sensing scenarios.

\textbf{2) Five-Billion-Pixels Dataset.}
The Five-Billion-Pixels (FBP) dataset~\cite{TONG-2023} is a high-resolution remote-sensing benchmark designed for large-scale fine-grained land-cover classification. It is constructed from Gaofen-2 satellite imagery with a spatial resolution of 4~m and covers more than $50{,}000$~km$^2$ across China, providing over five billion pixel-level annotations. The dataset adopts a highly detailed category system containing 24 land-cover/land-use classes, posing a substantial challenge to the ability of a model to transfer general semantic priors to complex fine-grained spatial patterns. In our experiments, FBP is treated as an unseen target domain. Its large semantic gap and considerable distribution shift are used to evaluate the cross-domain robustness of the proposed method. In particular, this dataset allows us to examine the effectiveness of entity-level geometric constraints and semantic advection refinement in suppressing out-of-domain noise and recovering fine-grained land-cover/land-use regions.

\textbf{3) Potsdam Dataset.}
The Potsdam dataset~\cite{ISPRSDATA-2014}, released by the International Society for Photogrammetry and Remote Sensing (ISPRS), is a widely used high-resolution aerial image benchmark. It has a spatial resolution of 0.05~m and contains four spectral bands, namely red, green, blue, and near-infrared. Potsdam includes densely distributed buildings, roads, vehicles, and low vegetation, thereby imposing stringent requirements on entity-boundary delineation and small-object recognition. In our experiments, Potsdam is regarded as an unseen target domain. Under the substantial spatial-resolution gap and cross-city landscape variation, this dataset is used to evaluate the ability of entity-level geometric constraints to preserve fine object boundaries, as well as the cross-domain generalization capability of semantic advection refinement in suppressing intra-class noise and enhancing the representation of small targets.

\textbf{4) Vaihingen Dataset.}
The Vaihingen dataset~\cite{ISPRSDATA-2014} is another high-resolution aerial remote-sensing benchmark released by ISPRS. It has a spatial resolution of 0.09~m and contains five major land-cover categories: \textit{impervious surface}, \textit{building}, \textit{low vegetation}, \textit{tree}, and \textit{car}. Unlike Potsdam, which provides standard RGB imagery, Vaihingen adopts a near-infrared--red--green (IR-R-G) band combination, resulting in a pronounced spectral distribution discrepancy. In our experiments, Vaihingen is treated as an unseen target domain to specifically assess the cross-domain few-shot segmentation capability of the proposed method under heterogeneous spectral-band configurations and substantial sensor-induced appearance shifts.

\textbf{5) iSAID Dataset.}
The iSAID dataset~\cite{iSAID-2019} is a large-scale aerial image instance segmentation benchmark derived from the DOTA dataset. It contains 2,806 multi-source high-resolution images with spatial resolutions ranging from 0.1~m to 4~m, mainly due to differences in acquisition platforms. The dataset covers 15 object categories, including \textit{ship}, \textit{harbor}, \textit{ground track field}, and \textit{bridge}. It is characterized by substantial scale variation, densely distributed objects, and numerous small targets with complex geometric structures. Because of its extreme multi-scale variation and crowded spatial layouts, iSAID constitutes a particularly challenging evaluation scenario. It is employed to assess the ability of entity-level few-shot segmentation and semantic advection refinement to achieve robust cross-domain representation and accurate segmentation under severe scale changes and dense-object interference.

\textbf{6) FLAIR \#1 Dataset.}
FLAIR \#1 \cite{FLAIR-2022} is a large-scale, high-resolution aerial-image semantic segmentation dataset developed by the French National Institute of Geographic and Forest Information (IGN). It covers diverse geographic regions of France and includes multi-temporal observations, with a spatial resolution of 0.2~m. The official benchmark maps the original labels to 13 land-cover/land-use classes. For our cross-domain few-shot segmentation experiments, we treat the \textit{other} class as background and use the remaining 12 classes as foreground classes. FLAIR \#1 exhibits substantial cross-region and cross-time distribution shifts. We use it to evaluate generalization and fine-grained segmentation under these shifts. 

\textbf{7) MiniFrance Dataset.}
MiniFrance \cite{MiniFrance-2020} is a large-scale, ultra-high-resolution aerial-image semantic segmentation dataset containing more than 2,000 RGB aerial images from 16 French cities and their surrounding areas. The images were acquired between 2012 and 2014. Pixel-level land-use and land-cover annotations were generated by rasterizing Urban Atlas 2012 data. We select 12 classes, including \textit{Urban fabric}, \textit{Industrial}, \textit{Mine}, \textit{Artificial non-agricultural vegetated areas} and et al., for cross-domain few-shot semantic segmentation evaluation. The dataset exhibits substantial variation across cities, climatic conditions, and urban and rural landscapes, making it suitable for evaluating cross-domain generalization under complex spatial distribution shifts. 

\subsection{Metrics}
To quantitatively evaluate the performance of the proposed method on the cross-domain few-shot segmentation task, we adopt the mean Intersection over Union (mIoU) as the primary evaluation metric. It is defined as the average Intersection over Union (IoU) over all foreground classes:
\begin{equation}
\mathrm{mIoU}
=
\frac{1}{C}
\sum_{c=1}^{C}
\frac{\mathrm{TP}(c)}
{\mathrm{TP}(c)+\mathrm{FP}(c)+\mathrm{FN}(c)},
\label{eq:miou}
\end{equation}
where $C$ denotes the number of foreground classes, and $\mathrm{TP}(c)$, $\mathrm{FP}(c)$, and $\mathrm{FN}(c)$ represent the numbers of true positive, false positive, and false negative pixels for class $c$, respectively.

To further evaluate the segmentation quality around object boundaries, we additionally report the Boundary Intersection over Union (BIoU)~\cite{BIoU_2021}. BIoU measures the intersection over union between the inner boundary regions extracted from the predicted and ground-truth masks of each class: 
\begin{equation}
\mathrm{BIoU}
=
\frac{1}{C}
\sum_{c=1}^{C}
\frac{\mathrm{TP}_{b}(c)}
{\mathrm{TP}_{b}(c)+\mathrm{FP}_{b}(c)+\mathrm{FN}_{b}(c)},
\label{eq:biou}
\end{equation}
where $\mathrm{TP}_{b}(c)$, $\mathrm{FP}_{b}(c)$, and $\mathrm{FN}_{b}(c)$ denote the numbers of true positive, false positive, and false negative pixels within the boundary region of class $c$, respectively. The boundary width is typically defined as a fixed proportion of the image diagonal. Following the standard setting, the default ratio is set to $0.02$ in all experiments.

\subsection{Implementation Details}
The proposed method is implemented based on PyTorch. All experiments are conducted on a workstation equipped with four NVIDIA GeForce RTX 3090 GPUs, each with 24~GB of memory. Since the proposed framework is training-free, no source-domain training, target-domain fine-tuning, or parameter updating is performed throughout the entire experiments. All parameters of SAM3 remain frozen during inference, and FP16 mixed-precision inference is adopted to improve computational efficiency. All input images are uniformly cropped into patches of $512\times512$ pixels. Following the standard few-shot segmentation protocol, both 1-shot and 5-shot settings are evaluated, where one or five support patches from the target domain are provided for each class to segment the corresponding query image. For all datasets, the text prompts are directly constructed from the official class names without any manual prompt engineering. We perform five random support-sampling runs with different random seeds and report the mean mIoU and standard deviation across runs.

The backbone configurations of our method and the compared methods are as follows. ELFSS-AR (ours) and ConceptBank both use SAM3 models with a PE-L+/14 visual backbone. FSSAM combines SAM2 and DINOv2~\cite{DINOv2_2024}, whose visual backbones are Hiera-S and ViT-B/14, respectively. PANet, PFENet, SSP, IFA, DR-Adapter, APM, and ICPD all use ResNet-50 as their visual backbone, while ICPD additionally incorporates RemoteCLIP~\cite{Remoteclip_2024} with a ViT-L/14 visual backbone. MaskCLIP, ClearCLIP, SCLIP, ResCLIP, and SegEarth-OV all employ CLIP models~\cite{CLIP_2021} with a ViT-L/14 visual backbone. Since the feature representations, module designs, and training or inference procedures of these methods are closely coupled with their respective backbones, uniformly replacing them with SAM3 requires corresponding architectural adaptations and strategy adjustments. These adjustments alter the original implementations and introduce additional variables, meaning that the compared methods no longer correspond to the original methods. Therefore, to maintain consistency with their original designs and implementations, we retain the respective backbones of the compared methods.

\section{Ablation and Sensitivity Analysis of Entity-Level Semantic Inference}

\begin{figure}[t]
\centering
\includegraphics[width=0.48\textwidth]{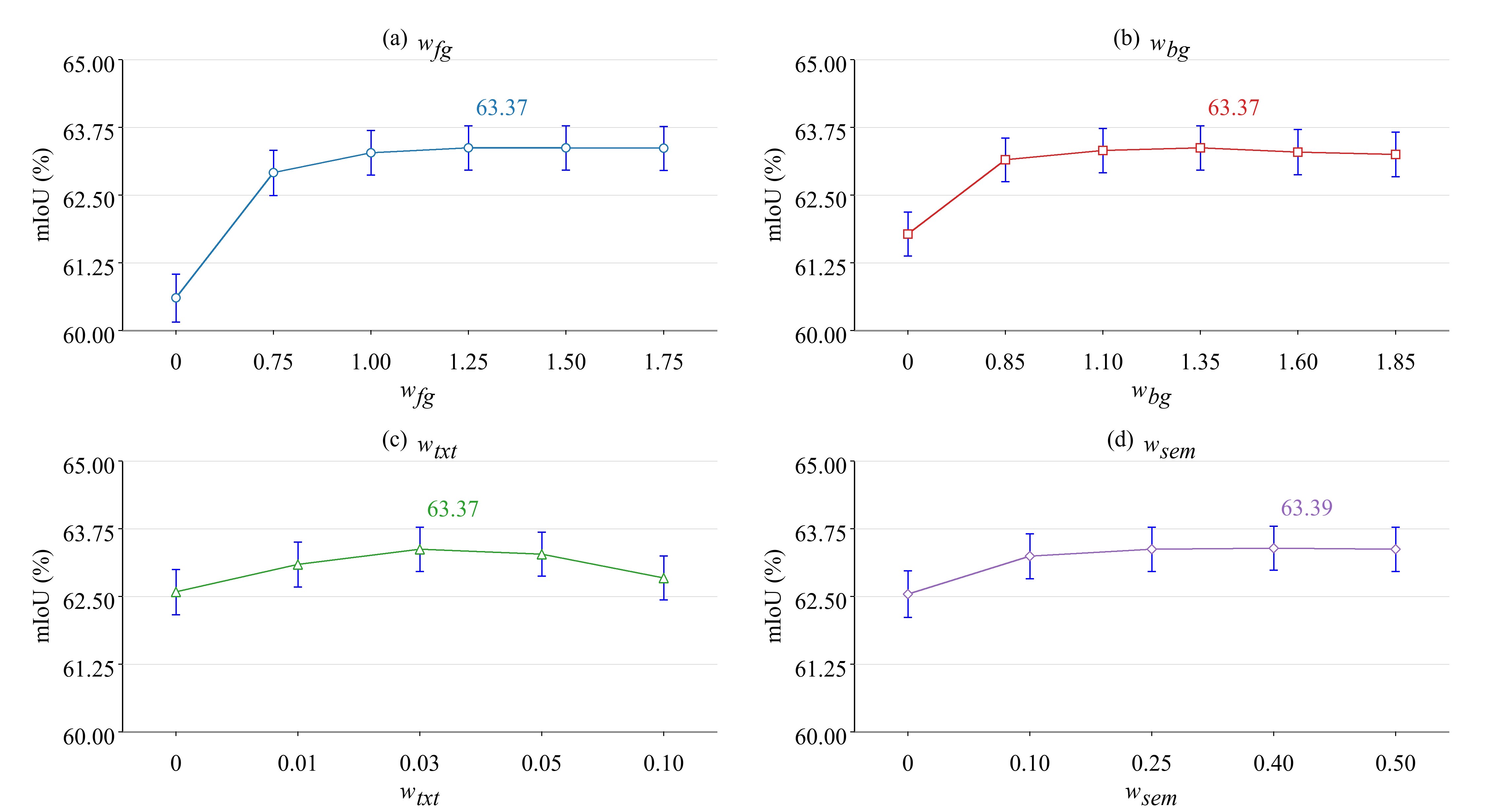}
\caption{Sensitivity analysis of fusion weights for entity-level semantic inference on the Vaihingen dataset under the 5-shot setting. Error bars show standard deviations scaled by $0.2$.}
\label{fig:parameter_entity_level_semantic_inference}
\end{figure}

Here, we analyze the contributions of individual scoring terms and the effects of their fusion weights in entity-level semantic inference. The fused score is defined as: 
\begin{equation}
\begin{aligned}
z_{i,c}={}& w_{\mathrm{fg}}\frac{\mu_{i,c}^{\mathrm{fg}}+1}{2}-w_{\mathrm{bg}}\frac{\mu_{i,c}^{\mathrm{bg}}+1}{2}+w_{\mathrm{txt}}\frac{\mu_{i,c}^{\mathrm{txt}}+1}{2}
\\
&+w_{\mathrm{sem}}\mu_{i,c}^{\mathrm{sem}}+w_{\mathrm{sam}}s_i^{\mathrm{sam}}.
\end{aligned}
\label{eq:fusion_score_suppl}
\end{equation}

The first four terms in Eq.~(\ref{eq:fusion_score_suppl}) are collectively referred to as the semantic scoring terms and directly determine the class label assigned to all pixels within the current entity primitive. The last term represents the entity quality score, which provides auxiliary information for downstream tasks. Fig.~\ref{fig:parameter_entity_level_semantic_inference} presents an analysis of the contributions of the semantic scoring terms and the sensitivity to their weights. All symbols are defined in the “Entity-Level Semantic Inference” section of the main text, so we omit their definitions here to avoid repetition. All results are reported as the mean mIoU over five random seeds under the 5-shot setting, with the remaining experimental settings kept fixed on the Vaihingen dataset.

\subsection{Contribution of Individual Scoring Terms}
\label{sec:contri_ind_scor_terms}

To evaluate the contribution of each semantic scoring term, we set its corresponding weight to zero while keeping the remaining settings unchanged and compare the results with the full configuration. The full model achieves an mIoU of 63.37\% in default settings which are detailed in the main text. As shown in Fig.~\ref{fig:parameter_entity_level_semantic_inference}, removing the foreground scoring term ($w_{\mathrm{fg}}=0$) decreases mIoU by 2.77 percentage points, indicating that the foreground response serves as the primary basis for entity-level class discrimination. Removing the background scoring term ($w_{\mathrm{bg}}=0$) decreases mIoU by 1.59 percentage points, demonstrating that the background response also provides complementary information for entity-level class discrimination. Removing the textual and semantic potential scoring terms ($w_{\mathrm{txt}}=0$ and $w_{\mathrm{sem}}=0$) decreases mIoU by 0.79 and 0.83 percentage points, respectively, indicating that both provide additional semantic cues for entity-level inference.

The entity quality scoring term $s_i^{\mathrm{sam}}$ adds the same score offset to the fused scores of all classes for a given entity primitive and therefore does not change its predicted class. Accordingly, our mIoU sensitivity analysis focuses on the remaining four fusion weights. Nevertheless, we retain this term as a complementary source of information for assessing primitive reliability. A higher entity quality score produced by SAM3 indicates that an entity primitive is more reliable. In terms of image content, such high-quality entity primitives often correspond to objects with clear boundaries, clear separation from the background, distinct textures and structures, and relatively uniform spectral values among their internal pixels. Incorporating this quality score through weighted fusion allows us to distinguish the segmentation confidence of pixels belonging to different entities that have the same semantic score but differ in entity quality, providing additional information for downstream tasks. 

\subsection{Sensitivity to Fusion Weights}

To analyze the model's sensitivity to fusion weights, we vary the foreground, background, textual, and semantic potential fusion weights individually while keeping the remaining default settings unchanged. As shown in Fig.~\ref{fig:parameter_entity_level_semantic_inference}, increasing the foreground weight $w_{\mathrm{fg}}$ from its default value of 1.25 to 1.75 decreases mIoU by only 0.01 percentage points, indicating relatively stable performance at the default and larger values. A similar trend is observed for the background weight $w_{\mathrm{bg}}$. When varying the text weight $w_{\mathrm{txt}}$, the highest mIoU of 63.37\% is obtained at its default value of 0.03, with performance tending to decline as the weight increases further. For the semantic potential field weight $w_{\mathrm{sem}}$, the mIoU reaches 63.39\% at 0.40, an improvement of 0.02 percentage points over the default configuration. Further increases in this weight lead to a slight performance decline. Overall, the model's performance begins to stabilize once this weight $w_{\mathrm{sem}}$ reaches its default value of 0.25.

\section{Theoretical Analysis of the Effectiveness of Semantic Advection Refinement}
ELFSS-AR extends conventional few-shot segmentation (FSS) from pixel-level prediction to entity-level prediction. Its ability to alleviate the noise and fragmentation inherent in pixel-level predictions is therefore straightforward. Consequently, the effectiveness analysis of ELFSS-AR mainly focuses on the semantic advection refinement module. In the main text, the effectiveness of semantic advection refinement has already been validated experimentally through the ablation studies. Here, we further provide a theoretical analysis of its effectiveness. For clarity and brevity, we present the theoretical analysis only for the query feature advection refinement process. The analysis for the multi-source response advection refinement process follows the same principles and is therefore omitted.

\textbf{Lemma 1}: Advection refinement transports features from the approximate upstream location $px-\eta \mathbf{v}_c^F(px)$ rather than generating new features at the current location $px$.

\textbf{Proof}: The semantic directional gate is defined in the main paper as
\begin{equation}
G_c^{F,(l)}=\mathbf{1}\!\left(d_c^{F,(l)}<0\right),
\label{eq:gate_suppl}
\end{equation}
where $\mathbf{1}(\cdot)$ denotes the indicator function, which equals $1$ if $d_{c}^{F,(l)}<0$, and $0$ otherwise. The intermediate feature after applying the semantic directional gate is 
\begin{equation}
\tilde{F}_{q}^{c,(l+1)}=F_{q}^{c,(l)}-\delta^F G_{c}^{F,(l)}\left(v_{c,x}^{F}D_xF_{q}^{c,(l)}+v_{c,y}^{F}D_yF_{q}^{c,(l)}\right).
\label{eq:feature_euler1}
\end{equation}

After the residual interpolation as follows:
\begin{equation}
\bar{F}_q^{c,(l+1)} = (1-\gamma^F)F_q^{c,(l)}+\gamma^F\tilde{F}_q^{c,(l+1)}, 
\label{eq:feature_update}
\end{equation}
we obtain the refined feature map: 
\begin{equation}
\begin{aligned}
\bar{F}_q^{c,(l+1)}=F_{q}^{c,(l)}-\eta G_c^{F,(l)}\Big(v_{c,x}^{F} D_xF_{q}^{c,(l)}+v_{c,y}^{F} D_yF_{q}^{c,(l)}\Big),
\end{aligned}
\label{eq:feature_advection}
\end{equation}
where $\eta = \gamma^F \delta^F > 0$ denotes the effective feature-advection step size after residual interpolation. When advection is inactive at a spatial location $px$, i.e., $G_c^{F,(l)}(px)=0$, we have $\bar{F}_{q}^{c,(l+1)}(px)=F_{q}^{c,(l)}(px)$. When advection is active, the feature at this location is updated as follows:

\begin{equation}
\begin{aligned}
\bar{F}_{q}^{c,(l+1)}(px)
=&F_{q}^{c,(l)}(px)-\eta \Big(
v_{c,x}^{F}(px)D_xF_{q}^{c,(l)}(px)\\
&+v_{c,y}^{F}(px)D_yF_{q}^{c,(l)}(px)
\Big),
\end{aligned}
\label{eq:active_feature_advection}
\end{equation}
This update (\ref{eq:active_feature_advection}) corresponds to an Euler discretization and provides a numerical approximation to the solution of the advection equation, which is given in the main text as
\begin{equation}
\frac{\partial F_q^c}{\partial t}+v_{c,x}^{F}\frac{\partial F_q^c}{\partial x}+v_{c,y}^{F}\frac{\partial F_q^c}{\partial y}=0.
\label{eq:feature_pde}
\end{equation}
Furthermore, applying a first-order Taylor expansion to the query feature gives: 
\begin{equation}
\begin{aligned}
&F_{q}^{c,(l)}\left(px-\eta \mathbf{v}_c^F(px)\right) \approx F_{q}^{c,(l)}(px)\\
&\qquad-\eta\left(v_{c,x}^F(px)D_xF_{q}^{c,(l)}(px)+v_{c,y}^F(px)D_yF_{q}^{c,(l)}(px)\right).
\end{aligned}
\label{eq:feature_taylor}
\end{equation}
Combining Eqs.~(\ref{eq:active_feature_advection}) and~(\ref{eq:feature_taylor}), we obtain $\bar{F}_{q}^{c,(l+1)}(px)
\approx F_{q}^{c,(l)}\left(px-\eta\mathbf{v}_c^F(px)\right)$. In this first-order sense, when advection is active at a spatial location $px$, the feature at $px$ can be interpreted as being transported from the approximate upstream location $px-\eta \mathbf{v}_c^F(px)$ rather than generated locally.

\textbf{Proposition 1}: The semantic directional gate adaptively selects the spatial locations for feature advection based on the directional derivative of the current feature semantic potential field along the fixed velocity field. It favors feature transport from locations with higher semantic confidence to locations with lower semantic confidence, thereby imposing a semantic directional constraint on feature advection. 

\textbf{Proof}: When the semantic gate as shown in (\ref{eq:gate_suppl}) is active, the following inequality at position $px$ holds: 
\begin{equation}
\begin{aligned}
&d_{c}^{F,(l)}(px)\\
&=v_{c,x}^{F}(px) D_xU_{c}^{F,(l)}(px)+v_{c,y}^{F}(px) D_yU_{c}^{F,(l)}(px) < 0.
\end{aligned}
\label{eq:feature_gate_suppl}
\end{equation}
Applying a first-order Taylor expansion to the current semantic potential field at the location $px-\eta \mathbf{v}_c^F(px)$, we obtain
\begin{equation}
\begin{aligned}
&U_c^{F,(l)}\left(px-\eta \mathbf{v}_c^F(px)\right)\\
&\approx U_c^{F,(l)}(px)\\
&\quad-\eta\left(v_{c,x}^F(px)D_xU_c^{F,(l)}(px)+v_{c,y}^F(px)D_yU_c^{F,(l)}(px)\right)\\
&=U_c^{F,(l)}(px)-\eta d_c^{F,(l)}(px).
\end{aligned}
\label{eq:potential_taylor}
\end{equation}

Since $\eta>0$ and $d_c^{F,(l)}(px)<0$, we have:
\begin{equation}
\left\{
\begin{aligned}
U_c^{F,(l)}\left(px-\eta\mathbf{v}_c^F(px)\right)
&\approx U_c^{F,(l)}(px)-\eta d_c^{F,(l)}(px),\\
-\eta d_c^{F,(l)}(px)&>0.
\end{aligned}
\right.
\label{eq:potential_increase}
\end{equation}

According to Lemma~1, in the local first-order sense, the feature at $px$ after each advection iteration can be interpreted as being transported from $px-\eta\mathbf{v}_c^F(px)$. Equation~(\ref{eq:potential_increase}) shows that the first-order approximation produces a higher semantic potential at this approximate upstream location $px-\eta\mathbf{v}_c^F(px)$ than at the current target location $px$. Thus, in this local first-order interpretation, the gate favors feature transport from locations with higher semantic potential to those with lower semantic potential. This transport direction is reasonable because a higher semantic potential indicates greater segmentation confidence. By allowing features from high-confidence locations to influence neighboring low-confidence locations, this mechanism aims to strengthen the feature representations at these low-confidence locations.

\end{document}